\documentclass{article}

\PassOptionsToPackage{numbers,compress}{natbib}
\usepackage[preprint]{neurips_2024}

\usepackage[utf8]{inputenc}
\usepackage[T1]{fontenc}
\usepackage{hyperref}
\usepackage{url}
\usepackage{booktabs}
\usepackage[flushleft]{threeparttable}
\usepackage{amsfonts}
\usepackage{amsmath}
\usepackage{amssymb}
\usepackage{nicefrac}
\usepackage{microtype}
\usepackage{xcolor}
\usepackage{graphicx}
\usepackage{multirow}
\usepackage{array}
\usepackage{longtable}
\usepackage{algorithm}
\usepackage{algpseudocode}
\usepackage{amsthm}
\usepackage[framemethod=TikZ]{mdframed}
\usepackage{enumitem}
\usepackage{orcidlink}   
\graphicspath{{figures/out/}{figures/}}

\hypersetup{
  pdftitle  = {IB2: A Protocol for Measuring Enterprise AI Systems
               by Serving Route, Not Model Identifier},
  pdfauthor = {Blake Stenstrom, Charangan Vasantharajan, Brian Sathianathan},
  hypertexnames = false,
  colorlinks = true,
  linkcolor  = black,
  citecolor  = black,
  urlcolor   = black,
  pdfborder  = {0 0 0},
}

\definecolor{phcolor}{HTML}{B3261E}
\definecolor{phbg}{HTML}{FDF3F2}

\newenvironment{phbox}[2]{%
  \begin{mdframed}[linecolor=phcolor,linewidth=0.8pt,backgroundcolor=phbg,
                   roundcorner=3pt,innertopmargin=6pt,innerbottommargin=6pt,
                   innerleftmargin=8pt,innerrightmargin=8pt,skipabove=6pt,skipbelow=6pt]
  {\color{phcolor}\sffamily\bfseries PLACEHOLDER #1 \quad\textnormal{\sffamily #2}}\par\vspace{2pt}
  \small\sffamily\color{phcolor}%
}{\end{mdframed}}

\makeatletter
\@ifundefined{submissioncheck}{}{%
}
\makeatother

\newtheorem{definition}{Definition}
\newtheorem{property}{Property}

\newcommand{\yes}{$\bullet$}
\newcommand{\pt}{$\circ$}
\newcommand{\no}{\text{--}}
\newcommand{\ibib}{\texorpdfstring{IB\textsuperscript{2}}{IB2}}

\title{\ibib{}: A Protocol for Measuring Enterprise AI Systems\\
by Serving Route, Not Model Identifier}

\author{%
  Blake Stenstrom\thanks{Corresponding author: \texttt{blake@iterate.ai}} \quad
  Charangan Vasantharajan~\orcidlink{0000-0001-7874-3881} \quad
  Brian Sathianathan \\
  Iterate.ai \\
  \texttt{\{blake, charangan, brian\}@iterate.ai}
}

\begin{document}

\maketitle

\begin{abstract}
Enterprises deploy systems, not checkpoints. Usable capability depends jointly
on weights, serving route, precision, output contract, and harness, yet all 18
audited benchmarks score advertised model identifiers. We treat this as
measurement error and give a protocol that makes it reportable. It has three
parts. A gold-blind \emph{capability-binding preflight} verifies that a route
can execute the evaluation contract before any task reaches it; a
\emph{reliability-inclusive} first-pass scoring rule keeps failure in the score
while keeping unsupported capability out; and adjudication is
\emph{structurally score-blind}. We call the protocol \ibib{} and release its
algorithms, classification tables, request contract, and manifest schemas. Its
reference instantiation, 128 locked tasks and 987 assertions over document,
spreadsheet, chart, tool and database work, stays sealed: the procedure is the
artifact, not the corpus. Across eleven systems, four results. Capability
availability is measurable: two complete single-route runs on identical weights
later failed distinct predicates of the finalized binding gate, while a third
passed that gate before a fresh run. The advertised identifier exposed
neither limit. Discrimination is not uniform: four of seven suites saturate
under a six-system band, with the spread almost entirely from governed database
work and multi-tab joins, so we report interval-backed resolution groups, not
ranks; two of the nominal five-label output's four cuts fail multiplicity
adjustment. Serving-arm choice moved one declared revision and precision from
77.38 to 82.54, paired interval $[0.11,10.60]$, though the arms differ in
access mode, harness generation, and the serving tool-call parser, and harness
generation is a property of our evaluator, not any endpoint. Excluding failed
responses from denominators changes the point ordering, so reliability
inclusion changes a conclusion, not its wording.
\end{abstract}

\section{Introduction}

\subsection{The evaluation gap: checkpoints versus deployed systems}

Every capability benchmark in our audit reports scores against model
identifiers. Enterprises buy
served endpoints. The gap between those two objects is not a rounding error. A
served route imposes its own completion ceiling, image limit, tool-call parser,
guided-decoding backend and numerical precision, and each of these can change
what a set of weights is observably able to do. A benchmark that never checks
the route cannot tell a model that failed from a route that refused, and
therefore cannot report which of the two its number describes. We claim that,
and only that. We do not claim that any published benchmark has in fact
misattributed a route limit to a model: we found no documented instance, and
establishing one would require per-route records that leaderboards do not
retain. The defect we address is an unreportable quantity, not a catalogue of
errors.

The problem is sharpest where enterprise work lives. Short question answering
exercises little of the serving envelope. A single task in the bank below may
require thirteen images in one request, selection from a twenty-five-tool
catalog, tool-result continuation across turns, a strict JSON output contract,
and a completion longer than 32{,}768 tokens, and routes serving the same
advertised model revision diverge on every one of those axes. The envelope in
Table~\ref{tab:contract} is a design choice of ours rather than a measured
workload population, and we hold no evidence that it is the modal enterprise
request, so we scope every route finding in this paper to \emph{this} contract.
What that scoping does not weaken is the structural point: whatever contract a
benchmark adopts, whether a route can execute it is a property of the route.

\subsection{A motivating failure}
\label{sec:motivating}

The retained lineage is sequential rather than a three-route prospective
preflight. DeepInfra FP8 (R11, 2026-07-24) and CoreWeave FP8 (R12,
2026-07-26) each ran all 128 locked tasks once and scored 26.12 and 62.25,
respectively. A later gold-blind capability audit found that DeepInfra rejected
the unchanged thirteen-image contract at its four-image route limit, and that
CoreWeave---despite advertising a 262{,}144-token completion maximum---stopped
at exactly 32{,}768 tokens with \texttt{finish\_reason=length}. The complete
ten-predicate gate, including that long-output predicate, was first attested on
2026-08-13, after R12. A third FP8 route, called Provider~P here, then passed
the full gate before its fresh 128-task replacement run. Its identity stays in
a signed private receipt and is withheld because the provider's terms prohibit
benchmarking (Section~\ref{sec:ethics}).

Same weights and same requested contract, but three separate whole runs: no
response was spliced across routes. The first two routes could execute enough
of the locked bank to produce reliability-inclusive scores, but could not meet
the full finalized envelope. Those two findings are retrospective relative to
R11/R12; only Provider~P was bound prospectively under the complete gate. A
harness that evaluates by model identifier has no place to preserve either
distinction. The claim is that route choice, capability limits, and chronology
must be part of the reported object.

\subsection{Research questions and contributions}

\textbf{RQ1 (validity)} Does route-level capability binding change the measured
capability profile and tier assignment of a system, relative to evaluating
against an advertised model identifier? \textbf{RQ2 (reliability)} How far does
a reliability-inclusive first-pass score diverge from an accuracy-only score,
and does that divergence change any tier assignment? \textbf{RQ3
(discriminative power)} Does the saturation criterion identify which suites
have stopped separating frontier systems, and what does it report here? RQ1 and
RQ2 are answered in Section~\ref{sec:ablations}, RQ3 in
Section~\ref{sec:suiteresults}, partly against our own interest.

The contribution is four-part. \textbf{C1} states the measured object:
Definition~\ref{def:sut} binds weights, route, precision, reasoning effort,
envelope, tool implementation, output contract and harness into one tuple, and
every score here is a score of that tuple. \textbf{C2} specifies four
algorithms with the tables they dispatch on: capability-binding preflight
(Algorithm~\ref{alg:bind}), gold-blind adjudication and recovery, fail-closed
comparability and resume, and suite-stratified task-set bootstrap. These were
implementation practice; here they are the contribution. We report two places
where the specification is ahead of what we ran, because for a protocol that
gap is itself a result. \textbf{C3} reports a route-availability outcome no
model-identifier benchmark can express, discloses whether each binding decision
was prospective or retrospective, and recomputes the cohort under each
convention the protocol rejects. \textbf{C4} is the instrument and cohort: a
128-task, 987-assertion enterprise bank and an eleven-system evaluation. It is
documented through released methodology, executable reference code, schemas,
synthetic conformance fixtures, and aggregate results derived from retained
signed evidence rather than through a public corpus.

\subsection{What the name refers to, and what this does not claim}
\label{sec:naming}
\label{sec:scope}

\ibib{} expands to \emph{Iterate Business Intelligence Benchmark}, which is
where the squared form comes from: the four initials are IB written twice. One
name does two jobs. \ibib{} is the protocol, contribution C1--C3, and it is
what is released. Its \emph{reference instantiation} is the seven suites, 128
locked tasks and 987 assertions of Section~\ref{sec:construction}, which is C4
and whose corpus is sealed. Every score carries a profile name, \ibib-7 Full or
\ibib-6 Core, and a profile is always a property of the instantiation. That
line has a cost we pay in Table~\ref{tab:main}: the Qwen3.6-27B row's route is
withheld, so by our own definition that row is under-identified. We retain it
rather than silently improve the cohort by dropping its weakest complete run,
mark it, and say plainly that it is the one row not reproducible in principle
by a third party.

\ibib{} measures a model plus a fixed harness on synthetic enterprise artifacts
through a pinned route. It does not measure a production deployment: there is
no retrieval index over a live store, no human in the loop, no multi-user
state, and the tool layer is a controlled sandbox. We do not claim run-to-run
variance estimates, since each configuration is run once against a locked bank.
We do not claim a human baseline, and therefore claim no procurement-grade
interpretation for any score. And we do not claim that spreadsheet reasoning,
document question answering, chart understanding, tool calling or text-to-SQL
is individually novel; all five have established benchmarks.

\section{Related work and positioning}
\label{sec:related}

\paragraph{Single-capability antecedents.} Every \ibib{} suite has one, and none
of them binds the serving route before scoring or reports capability
availability as an outcome. Spreadsheet reasoning has SpreadsheetBench and its
end-to-end successor \citep{ma2024spreadsheetbench,zhu2026spreadsheetbench2};
document understanding has DocVQA, MMLongBench-Doc and M3DocRAG
\citep{mathew2021docvqa,ma2024mmlongbenchdoc,cho2024m3docrag}; charts and
rendered tables have ChartQA, ChartQAPro and TableVQA-Bench
\citep{masry2022chartqa,masry2025chartqapro,kim2024tablevqa}; tool use has
ToolLLM and BFCL \citep{qin2023toolllm,patil2025bfcl}; text-to-SQL has
Spider~2.0 and BIRD \citep{lei2024spider2,li2023bird}. These works establish the
tasks, and we claim novelty for none of them.

\paragraph{Enterprise and agentic evaluation.} The closest work is
OfficeQA~Pro, which evaluates grounded reasoning over an archival collection of
U.S.\ Treasury documents combining narrative text with numeric tables
\citep{opsahlong2026officeqapro}. It is the direct comparator for our
selectable-text document suite and the strongest counter-evidence to any claim
that our bank is hard: it places frontier systems in a far lower band than we do
on nominally the same capability. The difference is instructive rather than
embarrassing. Its evidence is a collection in which the controlling facts must
first be located; ours is a bounded thirty-page packet in which they are
guaranteed present. Section~\ref{sec:saturation} treats that as evidence about
which construct each instrument measures. $\tau$-bench and $\tau^2$-bench
evaluate conversational agents under dual control, and their $\mathrm{pass}^k$
is prior art for reliability-inclusive scoring that we adopt rather than claim
\citep{yao2024taubench,barres2025tau2bench}. CRMArena-Pro covers business
scenarios with multi-table dependencies and overlaps our tool and database
suites \citep{huang2025crmarenapro}. GDPval grades economically valuable work
across 44 occupations, and differs from \ibib{} in one direction: it grades open-ended deliverables by expert human judgement rather than by
deterministic assertions against a typed contract \citep{patwardhan2025gdpval}.
Finch, MBABench and SpreadsheetBench~2 are the direct 2025--26 comparators for
S1 and S2 \citep{dong2025finch,yen2026mbabench,zhu2026spreadsheetbench2}.

\paragraph{Reporting discipline.} Reporting cost and latency beside accuracy so
that trade-offs stay visible is the design philosophy of HELM, which we adopt
wholesale for Table~\ref{tab:ops} \citep{liang2023helm}.
\citet{hua2026knowledgework} propose a four-field schema for represented
activity, tested setting, required work product and evaluated result; we adopt
it as the primary form of Table~\ref{tab:suites}. Best practices for rigorous
agentic benchmark construction have been consolidated into a checklist, against
which we claim conformance item by item in Appendix~\ref{app:abc}
\citep{zhu2025abc}. Private-curator evaluation carries documented risks, including
unfalsifiability by the reader \citep{bansal2025peeking}.
Section~\ref{sec:ethics} engages that risk directly instead of treating privacy
as a settled operational choice. Construct validity in language-model evaluation has
recently been given quantitative treatment \citep{kearns2026constructvalidity},
which bears on Section~\ref{sec:saturation}.

\paragraph{Where route binding comes from.}
\label{sec:mlperf}
Binding the serving stack before publishing a number is not new. It is the
organizing principle of MLPerf Inference, which applies to ML inference
submissions the standard systems-testing notion of a \emph{system under test}
that Definition~\ref{def:sut} uses \citep{reddi2020mlperf}. MLPerf
enforces the binding structurally through its closed and open divisions. A
closed-division submission fixes the model, the preprocessing and the quality
target, and reports the result against the disclosed hardware, engine, precision
and batching configuration rather than against a model name. Systems
benchmarking settled this a decade before capability benchmarking met it. What
MLPerf does not do is let the served configuration change the \emph{measured
capability} and then report that change as an outcome, because that is not the
question it asks. That is the transfer we claim, and it is narrow on purpose:
the discipline is standard in systems benchmarking and is not represented by
any capability benchmark in our 18-work audit, where the convention is still
to score an advertised model identifier. We adopt two adjacent literatures
rather than arguing with them. Backend and harness choice alone shift measured
accuracy for identical weights
\citep{pape2026silenthyperparameter}, which is why $H$ is a component of
Definition~\ref{def:sut} rather than infrastructure; and quantization effects
concentrate in multi-step arithmetic and exact numeric output
\citep{kurtic2025bf16}, which is the task profile of our spreadsheet and
database suites. We name vLLM and SGLang as experimental variables rather than
as infrastructure \citep{kwon2023vllm,zheng2024sglang}.

\paragraph{Positioning.}
\label{sec:positioning}
Table~\ref{tab:positioning} places \ibib{} against eighteen prior works on ten
dimensions. The first five columns describe \emph{what} is evaluated and, for
our row, characterize the reference instantiation of
Section~\ref{sec:construction}; the last five describe \emph{how} and
characterize the protocol of Section~\ref{sec:protocol}. That distinction
matters when reading the release column: we withhold the corpus and publish the
procedure. \ibib{} is not alone in any of the first
five, nor in most of the last five: $\tau^2$-bench keeps first-pass failure in
its primary metric, HELM reports cost and latency beside accuracy, and MLPerf
Inference binds the full serving stack before a number is published. Within
this audit, \emph{Route bound} is the only column in which it stands alone
among \emph{capability} benchmarks, and we invite correction on that point
specifically. The
table carries no column for score-blind adjudication. One absent marker would
have to cover two different things: a benchmark that permits score-visible
recovery, and a benchmark that specifies no adjudication step at all. That makes
the column uninformative. The property itself is real and is stated in
Section~\ref{sec:adjudication}: of the eighteen comparators we audited, none
specifies that recovery decisions are taken without visibility of correctness,
and the row-by-row evidence is released as
\texttt{paper/data/positioning\_audit.csv}. Two columns are where \ibib{} is
weakest. The corpus is not public, a cost we account for in
Section~\ref{sec:ethics}, and there is no human baseline, which limits the
business interpretation of any absolute score.

\begin{table}[t]
\caption{Positioning against prior benchmarks. \yes{} present, \pt{} partial,
\no{} absent. The first five columns describe what is evaluated, the last five
how. \emph{Route bound}: the serving route is verified against the evaluation
contract before scoring. \emph{Reliab.\ in score}: first-pass failures remain in
the primary metric. \emph{Cost/lat.}: measured inference cost and latency
reported alongside accuracy (\pt{} if only one is reported). $^{\ast}$MLPerf binds the full serving stack, but
for throughput and latency at a fixed accuracy target rather than for measured
capability; see Section~\ref{sec:mlperf}. Cell values for prior work are
audited against each cited primary paper; the dated row-by-row evidence ledger
is released as \texttt{paper/data/positioning\_audit.csv}.}
\label{tab:positioning}
\centering
\scriptsize
\setlength{\tabcolsep}{2.6pt}
\begin{tabular}{@{}l ccccc ccccc@{}}
\toprule
Evaluation & Docs & Sheets & Charts & Tools & Gov.\ DB
& Corpus & Route & Reliab.\ & Cost/ & Human \\
 &  &  &  &  &  & public & bound & in score & lat. & baseline \\
\midrule
GPQA \citep{rein2023gpqa}                     & \no & \no & \no & \no & \no & \yes & \no & \no & \no & \yes \\
FinanceBench \citep{islam2023financebench}    & \yes& \no & \no & \no & \no & \yes & \no & \no & \no & \pt \\
SpreadsheetBench \citep{ma2024spreadsheetbench}& \no & \yes& \no & \no & \no & \yes & \no & \no & \no & \yes \\
SpreadsheetBench~2 \citep{zhu2026spreadsheetbench2} & \no & \yes & \no & \pt & \no & \yes & \no & \no & \no & \no \\
DocVQA \citep{mathew2021docvqa}               & \yes& \no & \no & \no & \no & \yes & \no & \no & \no & \yes \\
MMLongBench-Doc \citep{ma2024mmlongbenchdoc}  & \yes& \no & \pt & \no & \no & \yes & \no & \no & \no & \yes \\
TableVQA \citep{kim2024tablevqa}              & \pt & \no & \no & \no & \no & \yes & \no & \no & \no & \no \\
ChartQAPro \citep{masry2025chartqapro}        & \no & \no & \yes& \no & \no & \yes & \no & \no & \no & \yes \\
ToolLLM \citep{qin2023toolllm}                & \no & \no & \no & \yes& \no & \yes & \no & \no & \no & \no \\
BFCL \citep{patil2025bfcl}                    & \no & \no & \no & \yes& \no & \yes & \no & \pt & \no & \no \\
$\tau^2$-bench \citep{barres2025tau2bench}    & \no & \no & \no & \yes& \no & \yes & \no & \yes& \no & \no \\
CRMArena-Pro \citep{huang2025crmarenapro}     & \pt & \no & \no & \yes& \pt & \yes & \no & \pt & \pt & \no \\
Spider 2.0 \citep{lei2024spider2}             & \pt & \no & \no & \pt & \yes& \yes & \no & \no & \pt & \no \\
BIRD \citep{li2023bird}                       & \no & \no & \no & \no & \yes& \yes & \no & \no & \no & \yes \\
OfficeQA Pro \citep{opsahlong2026officeqapro}& \yes& \pt & \pt & \no & \no & \yes & \no & \no & \yes & \yes \\
GDPval \citep{patwardhan2025gdpval}               & \yes& \yes& \pt & \pt & \no & \pt & \no & \no & \no & \yes \\
HELM \citep{liang2023helm}                    & \pt & \no & \no & \no & \no & \yes & \no & \pt & \yes& \no \\
MLPerf Inference \citep{reddi2020mlperf}      & \no & \no & \no & \no & \no & \yes & \yes$^{\ast}$ & \no & \pt& \no \\
\midrule
\textbf{\ibib{} (this work)}                  & \yes& \yes& \yes& \yes& \yes& \no & \yes& \yes& \yes& \no \\
\bottomrule
\end{tabular}
\end{table}

\section{Problem formulation}

\subsection{The system under test}
\label{sec:sut}

\begin{definition}[System under test]
\label{def:sut}
A system under test is the tuple
$S = \langle m,\ r,\ p,\ e,\ \Omega,\ \mathcal{T},\ \mathcal{C},\ H \rangle$
where $m$ is the model identity including any pinned upstream revision, $r$ the
serving route including provider and disabled fallbacks, $p$ the served
numerical precision, $e$ the reasoning-effort setting, $\Omega$ the observed
context, completion, image and tool-call envelope, $\mathcal{T}$ the tool
implementation, $\mathcal{C}$ the output contract enforced on the wire, and $H$
the evaluation harness contract. A \emph{composed} system is a finite set of
such tuples with a routing function $\sigma$ mapping each suite to exactly one
of them, declared before binding; every component binds independently and the
composed system is reported under one identity only if all of them bind.
\end{definition}

The GLM entry routes text and tool suites to GLM-5.2 and visual suites to
GLM-5V-Turbo, which the single-tuple form cannot express. Two consequences are load-bearing for the rest of the paper.

\begin{property}[Non-substitutability]
\label{prop:nonsub}
Two systems differing only in $r$ are distinct systems under test. A score for
one does not transfer to the other, and their results may not be combined into
a single row.
\end{property}

\begin{property}[Asymmetric availability]
\label{prop:asym}
For closed-weight models no configuration exists in which $r$, $p$ and $\Omega$
are \emph{jointly} under the evaluator's control. A model-only score is
therefore not epistemically available for such systems, and cross-frontier
comparison between open- and closed-weight systems is possible only at the
level of $S$.
\end{property}

Property~\ref{prop:asym} is the strongest available justification for
system-level evaluation and is why we offer no model-only leaderboard: there is
no model-only number to offer for half the cohort, and for the other half such
a number would be a reference serving configuration reported under a model's
name. Section~\ref{sec:results} states the sense in which $H$ is identified for
runs that predate the current harness contract.

\subsection{The evaluation contract}
\label{sec:contract}

Table~\ref{tab:contract} states the request envelope every system must satisfy.
It fixes $\Omega$ and $\mathcal{C}$, two of the eight components of the system
under test, so a system is not identified until both are pinned. The contract is also the
object Algorithm~\ref{alg:bind} tests, the object Section~\ref{sec:ablations}
varies, and the reason two routes in Section~\ref{sec:motivating}
later failed the finalized gate. An evaluation that does not publish
its contract cannot report contract executability as a result. The particular
values below are ours and are scoped accordingly (Section~\ref{sec:scope}); the
requirement that some contract be published and tested is the protocol's.

\begin{table}[t]
\caption{The evaluation contract. A route is admitted only if it executes every
row. SQL limits are bytes where stated, not character counts.}
\label{tab:contract}
\centering
\small
\begin{tabular}{@{}lp{3.7cm}p{5.4cm}@{}}
\toprule
Dimension & Requirement & Enforcement \\
\midrule
Images per request   & 13, byte-identical & rejected by 4-image routes; probe 4 \\
Tool catalog         & 25 tools at binding & selection verified live; probe 5 \\
Tool calls per task  & $\leq 12$, one per turn & evaluator-owned; provider parallelism cannot bypass \\
Turn structure       & final-answer turn required & evaluator state machine \\
Completion budget    & 65{,}536 requested; ${>}32{,}768$ observed & \texttt{finish\_reason=length} fails the probe \\
Output               & strict JSON schema on the wire & revalidated against the evaluator contract \\
SQL bounds           & 65{,}536-byte SQL; 100 projected columns; 200 rows; 16{,}384-byte cells; 262{,}144-byte response; 120 s & governed read-only executor \\
Telemetry            & usage, cost, reasoning replay & missing usage places cost on hold \\
Retention            & zero-data-retention eligible; collection denied & route ineligible otherwise \\
Concurrency          & 1, serial task issue & harness \\
\bottomrule
\end{tabular}
\end{table}

\subsection{Estimand, population, and what a score licenses}
\label{sec:estimand}

The estimand is the performance of one fully specified system $S$ on the locked
bank $\mathcal{B}$. It is tempting to declare $\mathcal{B}$ the population
outright and to decline $p$-values on the ground that a population has no
sampling distribution. That is not consistent with what
Algorithm~\ref{alg:bootstrap} does. The algorithm constructs a resampling
distribution over $\mathcal{B}$, and Section~\ref{sec:tiers} uses that
distribution, subject to the auditability bounds in Section~\ref{sec:sensitivity},
to license ordinal claims. We state the frame the machinery actually assumes rather than
the one that sounds more conservative.

Two estimands are in play and they are not the same object. The \emph{reported}
score is a census, the equal-suite mean of $S$'s task scores over all of
$\mathcal{B}$ with no sampling error, which is why every number in
Table~\ref{tab:main} is printed without a standard error. The \emph{interval} is
a sensitivity analysis against a different question: would this ordering hold on
a comparably constructed bank? Answering that requires treating the
$n_k$ tasks of suite $k$ as exchangeable draws from the suite-specific
generative process of Section~\ref{sec:construction}: the scenario families, the
distractor and decoy controls, and the admission gate that produced them. Under
that frame $\mathcal{B}$ is one realization of the process, the resampling
distribution is over alternative realizations, and ``excludes zero'' means the
difference is not an artifact of which tasks the process happened to emit. It is
not a claim about run-to-run variance, and not a claim about a population of
enterprise tasks in the world; the generative process is ours, and
Section~\ref{sec:scope} says what that does not license.

Three consequences follow. We do not report $p$-values: a sampling distribution
does exist, but the estimand is per-system, and 55 marginal $p$-values over a
multiplicity structure we do not control would mislead more than the tier
presentation does. We do not estimate run-to-run variance, because $n{=}1$ per
configuration and the resampling unit is the task, not the run. And a score
licenses an ordinal claim only where the relevant paired interval excludes zero;
where it does not, we report a shared resolution group.

\subsection{Work representation}
\label{sec:workrep}

A suite name such as ``charts'' does not say what work is represented, under
what conditions, or what counts as done. Table~\ref{tab:suites} therefore states every suite in the four-field schema of
\citet{hua2026knowledgework}. Its S3 and S4 rows are what let a reader judge the
construct question raised in Section~\ref{sec:saturation}.

\begin{table}[t]
\caption{Suite composition, stated in the four-field work-representation schema
of \citet{hua2026knowledgework}. Per-suite assertion counts and the realized
difficulty distribution are in Appendix~\ref{app:assertions}.}
\label{tab:suites}
\centering
\footnotesize
\setlength{\tabcolsep}{3.5pt}
\begin{tabular}{@{}lp{2.15cm}p{2.95cm}p{2.35cm}p{2.95cm}r@{}}
\toprule
 & Represented activity & Tested setting & Required work product & Evaluated result & Tasks \\
\midrule
S1 & Answer a business question from one worksheet
   & Synthetic single-tab workbook; ordered sheet record with source row numbers
   & Typed answer with cited source ranges
   & Value correctness under tolerance; citation of the controlling range & 10 \\
\addlinespace
S2 & Reconcile figures across worksheets
   & Multi-tab workbook; stale caches, effective dates, duplicate rows
   & Typed answer with join provenance
   & Join correctness; temporal rule application; duplicate resolution & 22 \\
\addlinespace
S3 & Extract and reconcile evidence from a document packet
   & 30-page selectable-text PDF; non-adjacent evidence, decoy tables
   & Typed answer with page-level citations
   & Value correctness; whether the cited page is the controlling one & 22 \\
\addlinespace
S4 & Read a table that exists only as an image
   & Image-only pages at 180 DPI; merged headers, footnotes; no extracted text supplied
   & Typed answer with cell references
   & Transcription fidelity; correctness of the downstream calculation & 22 \\
\addlinespace
S5 & Interpret a chart, organizational structure, schedule, or risk matrix
   & Rendered figures; dual axes, critical-chain schedules, architecture paths
   & Typed answer naming the visual element relied on
   & Read-off accuracy; structural inference over the depicted relation & 22 \\
\addlinespace
S6 & Execute a bounded enterprise workflow
   & Stateful sandbox; per-task tool subset, one call per turn, $\leq 12$ calls
   & Ordered tool calls and a final typed answer
   & Selection, order, arguments, side-effect safety, audit trail, final answer & 18 \\
\addlinespace
S7 & Answer a governed question spanning a database and documents
   & Read-only SQL over $10^7$ rows and a 1{,}000-column relation; certified term
     dictionary among decoys; PRAGMA blocked
   & SQL queries and a typed business answer
   & Schema-discovery credit; query correctness; final business answer & 12 \\
\midrule
\multicolumn{5}{@{}l}{\textbf{\ibib-7 Full}, 987 deterministic assertions} & \textbf{128} \\
\bottomrule
\end{tabular}
\end{table}

\section{Evaluation protocol}
\label{sec:protocol}

\begin{figure}[t]
\centering
\includegraphics[width=\textwidth]{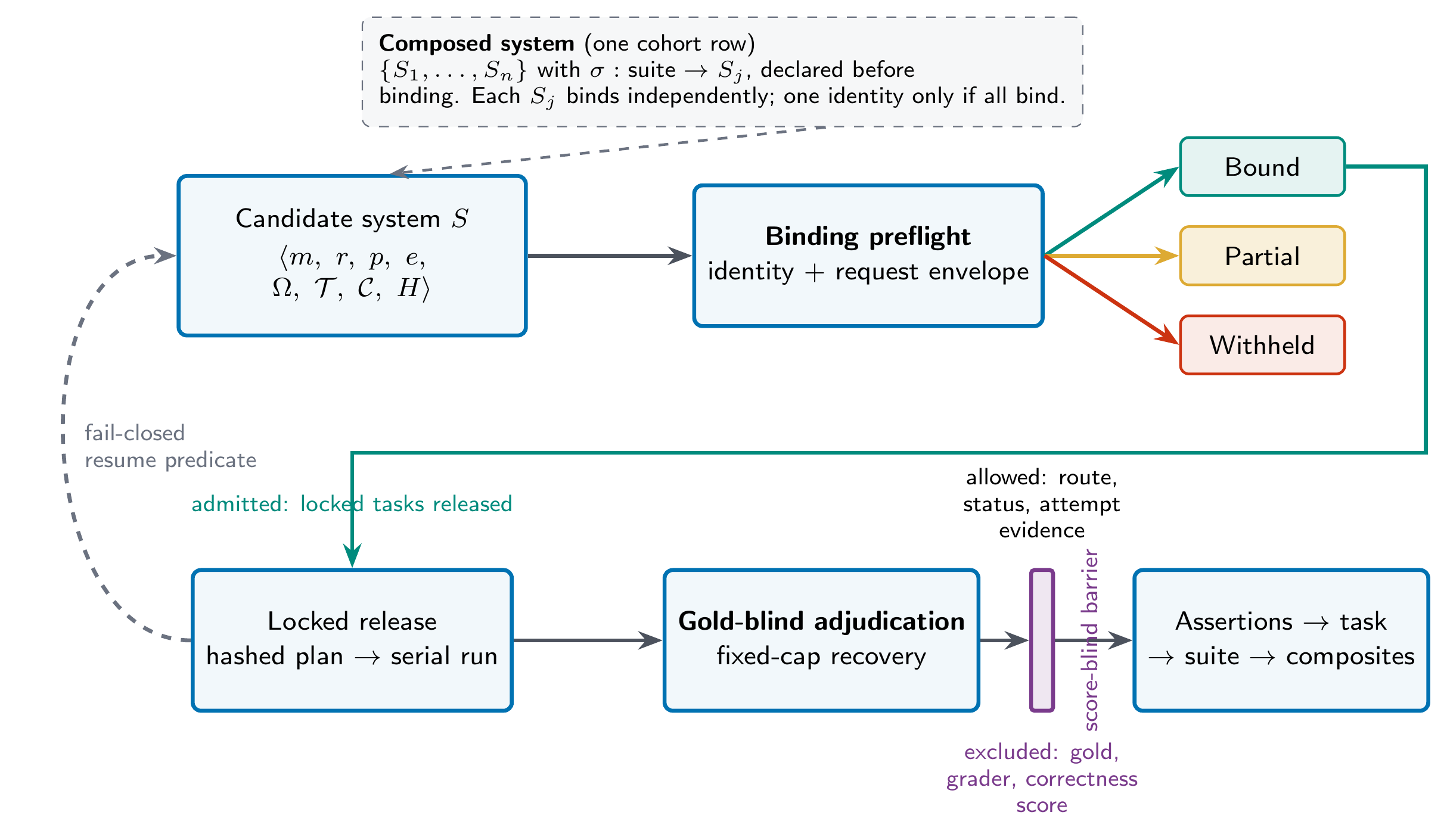}
\caption{The \ibib{} protocol. Binding has three explicit dispositions before a
locked task is released. Adjudication receives only the allowlisted projection;
gold, grader output, and correctness scores remain outside the recovery
selector. The dashed edge is the fail-closed resume predicate.}
\label{fig:protocol}
\end{figure}

Figure~\ref{fig:protocol} maps the protocol: binding, locked release under a
hashed plan, gold-blind adjudication of every non-OK response, and the
score-blind barrier the whole protocol rests on. 

\subsection{Harness neutrality and run-plan hashing}

Every system receives the same public task semantics, evidence bytes, task
order, tool budgets, and SQL boundary. Provider-native envelopes differ only
where an API requires a different wrapper, and the residual syntactic difference
is retained in the system manifest rather than claimed away. For each task, the
run plan hashes the public task, the rendered request, the system configuration,
the corpus manifest, and the material harness contract.

Conformance tests establish semantic and evidence equivalence, not that
vendors allocate the same hidden computation or run identical safety, OCR or
serving stacks. Those differences are properties \ibib{} measures, not
confounds it removes.

\subsection{Binding: an advertised identifier is not a capability}
\label{sec:binding}

An advertised model identifier or context window does not prove that a served
route can execute the evaluation contract. Algorithm~\ref{alg:bind} states the
gate: ten predicates, each carrying a class and a lane set, returning one of
\textsc{Bound}, \textsc{PartialCoverage} or \textsc{Withheld} together with the
class of the failure. A route-level incompatibility observed \emph{at binding}
is never converted into a model score of zero, and successful tasks from an
incompatible provider are never spliced into a nominally single-route result. Binding-time and run-time
failures differ. Nothing has been measured at binding, so there is nothing a
zero could describe. A route that truncates a locked task after admission has
produced a real failure of the system a customer would deploy, and
Algorithm~\ref{alg:adjudicate} scores it. Neither ever attributes the outcome
to the weights. This is the normative procedure; the reference lineage did not
apply its complete predicate set prospectively to R11 or R12. The dated
attestation in \texttt{paper/data/binding\_predicate\_freeze.csv} places the
first complete-set receipt at 2026-08-13 23:53:56 UTC, after R12 and before the
Provider~P run. We therefore treat R11/R12 as complete calibration runs later
audited against the finalized gate, not as preflight rejections.

Two properties of the gate are findings rather than description. First,
availability is two statistics, not one: predicates 1--7 ask whether the route
\emph{can} do the work and 8--10 whether it \emph{will disclose} what it did, so
a third party running this protocol should report capability availability and
governance availability separately. Both failures in this cohort are
capability-class, so nothing here turns on it. Second, the
\textsc{PartialCoverage} branch is unreachable in v0.13.2. It fires only when
the failing predicates touch S4 and nothing else, but no predicate in
Table~\ref{tab:probes} has that lane: the only vision predicate is probe~4,
whose lane is $\{\text{S4},\text{S5}\}$, and S5 is in Core. The cohort confirms
it. DeepInfra FP8 failed exactly probe~4, thirteen images against a four-image
route limit, and was dispositioned \textsc{Withheld} rather than
\textsc{PartialCoverage}: the correct outcome under the lane mapping, and the
wrong one under the intent Section~\ref{sec:suites} states. Making the branch
reachable needs a predicate whose lane is S4 alone. Probe~4 tests whether a
route accepts an image, which is a different question from whether it can read
one.

\begin{algorithm}[t]
\caption{Provider-capability binding preflight. Executed before any locked task
is released to a candidate route. The receipt is content-addressed and asserts
that no benchmark task, gold answer, or score was loaded or consulted. Each
predicate carries a class and a lane set, both published in
Table~\ref{tab:probes}, and the disposition returns the class of the failure
(Section~\ref{sec:binding}).}
\label{alg:bind}
\begin{algorithmic}[1]
\Require candidate route $r$, model identity $m$, evaluation contract
  $\mathcal{K} = \langle \mathcal{C}, \mathcal{T}, \Omega^{\ast} \rangle$
\Ensure disposition $d \in \{\textsc{Bound}, \textsc{Withheld},
  \textsc{PartialCoverage}\}$, failure class
  $f \in \{\textsc{None}, \textsc{Capability}, \textsc{Governance},
  \textsc{Mixed}\}$, and a content-addressed receipt $\rho$
\State $P \gets$ the ordered predicate set of Table~\ref{tab:probes}; each
  $\pi \in P$ carries a class $\operatorname{cls}(\pi) \in
  \{\textsc{Capability}, \textsc{Governance}\}$ and a lane set
  $\operatorname{lane}(\pi) \subseteq \{\text{S1},\ldots,\text{S7}\}$
\State $\Pi \gets \emptyset$; $\pi_{\ast} \gets \bot$
  \Comment{$\pi_{\ast}$: the predicate that raised, if any}
\ForAll{predicate $\pi \in P$ in table order}
  \If{$\pi_{\ast} \neq \bot$}
    \Comment{no further provider turn is issued after an exception}
    \State $t \gets \textsc{Fail}$ \textbf{if}
      $\operatorname{lane}(\pi) \cap \operatorname{lane}(\pi_{\ast}) \neq \emptyset$
      \textbf{else} $\textsc{Unobserved}$
    \State $\Pi \gets \Pi \cup \{(\pi, t, \bot)\}$; \textbf{continue}
  \EndIf
  \State evaluate $\pi$ from route metadata, fixed routing controls, or the
    ordered synthetic cases \textsc{Ordinary}, \textsc{Vision},
    \textsc{ToolStart}, \textsc{ToolContinue}, and \textsc{LongOutput}
  \State $\Pi \gets \Pi \cup \{(\pi, \textsc{Pass}/\textsc{Fail},
    \text{observed envelope})\}$
  \If{the request raised before a provider envelope was available}
    \State record the exception; overwrite $\pi$'s status with \textsc{Fail};
      $\pi_{\ast} \gets \pi$
  \EndIf
\EndFor
\State $F \gets \{\pi : \textsc{Fail}\}$;\quad
  $U \gets \{\pi : \textsc{Unobserved}\}$;\quad
  $L \gets \bigcup_{\pi \in F} \operatorname{lane}(\pi)$
\If{$F = \emptyset$ \textbf{and} $U = \emptyset$}
  \State $d \gets \textsc{Bound}$; $f \gets \textsc{None}$
  \Comment{route admitted; $\Omega$ recorded as observed, not as advertised}
\ElsIf{$L = \{\text{S4}\}$ \textbf{and} $U = \emptyset$}
  \State $d \gets \textsc{PartialCoverage}$; $f \gets \textsc{Capability}$
  \Comment{the lane is reported as uncovered; it is \emph{not} scored zero}
\Else
  \State $d \gets \textsc{Withheld}$
  \Comment{no locked task is released to $r$}
  \State $f \gets \textsc{Capability}$ if every $\pi \in F \cup U$ is
    \textsc{Capability}; \textsc{Governance} if every $\pi \in F \cup U$ is
    \textsc{Governance}; \textsc{Mixed} otherwise
  \Comment{an unobserved predicate carries its class, so a route that was
    never asked a governance question is not a clean capability failure}
\EndIf
\State $u \gets \langle$m, r, $\mathcal{K}$, $\Pi$, d, f, configuration and
  source hashes, no-benchmark assertions$\rangle$
\State $\rho.\texttt{receipt\_sha256} \gets
  \operatorname{SHA256}(\operatorname{JCS}(u))$
  \Comment{UTF-8 JSON, sorted keys, compact separators; hash field omitted}
\State \Return $(d, f, \rho)$
\end{algorithmic}
\end{algorithm}

\begin{table}[!tp]
\caption{The ten binding predicates, in evaluation order, with the class and
lane set Algorithm~\ref{alg:bind} dispatches on. They are implemented by one
catalog/configuration gate and five provider turns; a single turn may satisfy
several predicates. Every provider turn must also return the pinned model and
upstream route. \emph{Class} separates predicates that test whether the route
can do the work from predicates that test whether it will disclose what it did;
Algorithm~\ref{alg:bind} returns the class so that the two never appear as one
availability statistic. \emph{Lane} is the suite set whose executability the
predicate gates, and is the mapping the \textsc{PartialCoverage} branch reads.
A hard request error fails closed for predicates sharing a lane with the
raising predicate and marks lane-disjoint downstream predicates
\textsc{Unobserved}, which is not \textsc{Bound}-eligible either.}
\label{tab:probes}
\centering
\footnotesize
\setlength{\tabcolsep}{4pt}
\begin{tabular}{@{}r>{\raggedright\arraybackslash}p{2.9cm}ll>{\raggedright\arraybackslash}p{5.5cm}@{}}
\toprule
\# & Probe & Class & Lane & Pass predicate \\
\midrule
1  & Route and parameter identity & capability & all & exactly one catalog endpoint matches the pinned route/provider; returned model/provider match it; fallbacks are disabled; required parameters are advertised \\
2  & Context and requested cap & capability & all & catalog context equals the pinned value and the ordinary request carries the frozen completion ceiling \\
3  & Structured output & capability & all & ordinary turn carries strict JSON schema, returns the required envelope, and ends with \texttt{stop} \\
4  & Image envelope & capability & S4, S5 & 13 images are present on the wire with byte hashes unchanged; the vision response is contract-valid and ends with \texttt{stop} \\
5  & Tool catalog and selection & capability & S6 & exactly 25 tools are declared; the model emits exactly one call to the designated synthetic tool with the designated arguments \\
6  & Tool-result continuation & capability & S6, S7 & the second turn receives the tool result, replays provider reasoning exactly, returns the required final JSON, and ends with \texttt{stop} \\
7  & Completion ceiling & capability & all & the frozen-cap long turn returns at least 32{,}769 provider-reported completion tokens and ends with \texttt{stop}, not \texttt{length} \\
8  & Reasoning telemetry & governance & all & at least one provider turn contains non-empty reasoning or a positive reasoning-token count; the tool turn satisfies exact replay \\
9  & Usage and cost telemetry & governance & all & every provider turn is usage-bearing SSE with integer prompt/completion counts, at least one stream event, and numeric per-response cost \\
10 & Retention controls & governance & all & the bound routing configuration requires ZDR and sets data collection to \texttt{deny} \\
\midrule
\multicolumn{5}{@{}l}{\emph{Named as missing; not in the v0.13.2 gate, both freeze requirements (Section~\ref{sec:frozen})}} \\
-- & Strict-JSON adherence under reasoning & capability & all & the final assistant message parses as a bare schema-conformant object, no prose or fence, on a turn that also emits a non-empty reasoning trace (Section~\ref{sec:probegap}) \\
-- & OCR capability, distinct from image acceptance & capability & S4 & an image-only table is transcribed to the declared cell schema, on a route that has already passed probe~4 \\
\bottomrule
\end{tabular}
\end{table}

\subsection{Adjudication: recovery decisions that cannot see correctness}
\label{sec:adjudication}

Every non-OK response enters an adjudication ledger before recovery eligibility
is decided. What distinguishes this from a policy of merely ignoring scores is
structural: the implementation projects each result onto an explicit allowlist,
so the object the recovery selector reads is a type from which score, grader
output and gold reference are \emph{absent} rather than unread. The projection
does retain \texttt{status}, which is itself a classification of the response,
so the adjudicator has necessarily parsed what came back and knows a response
was malformed, empty or envelope-violating. What it cannot reach is any gold
reference, any grader output and any score. That is the invariant that matters
for p-hacking: correctness is what a biased selector would need in order to
retry selectively, and correctness is exactly what is unreachable. A
contract-valid but incorrect answer is therefore ineligible for every recovery
branch, and no reachable field distinguishes it from a correct one.
Algorithm~\ref{alg:adjudicate} and its decision table, Table~\ref{tab:classify},
are in Appendix~\ref{app:algorithms}.

\begin{algorithm}[t]
\caption{Gold-blind adjudication and recovery. $\pi_{\text{allow}}$ is a typed
projection, not a filter applied at read time. Stage~0 is the request-level
retry that runs inside the harness before a result record exists. It is a
named step here because whether a transport fault enters a primary score
depends on it. $\textsc{Classify}$ is
the decision table of Table~\ref{tab:classify}, evaluated in table order with
first match winning.}
\label{alg:adjudicate}
\begin{algorithmic}[1]
\Statex \textbf{Stage 0 (harness, before a result record exists).}
  A request that fails in transport is reissued byte-identically at most
  $c_{\text{req}} = 3$ times (four total request attempts, as fixed by
  \texttt{max\_attempts=4} in every signed run plan). If a reissue returns a provider
  envelope, that envelope is the result and no failure is recorded: the task
  never enters the taxonomy of Table~\ref{tab:failures} and the primary score
  is unaffected. If the cap is exhausted, a result record with
  $\texttt{status} \neq \textsc{Ok}$ is written and Stage~1 begins. Stage~0 is
  gold-blind by construction: it runs before any scoring path is reachable.
\Statex\hrulefill
\Require raw result record $x$ with status $\neq \textsc{Ok}$
\Ensure $\delta \in \{\textsc{Score}, \textsc{Recover}, \textsc{Retry},
  \textsc{Coverage}, \textsc{Hold}\}$
\State $\hat{x} \gets \pi_{\text{allow}}(x)$, where the typed projection
  $\pi_{\text{allow}}$ retains only
\Statex \quad $\langle$run id, task id, suite, status, coverage state,
  public hashes, redacted error, sanitized envelope summary$\rangle$
\State \textbf{invariant:} $\hat{x}$ contains no gold reference, no grader
  output, and no score; \texttt{status} is a response classification and carries
  no correctness information
\State $c \gets \textsc{Classify}(\hat{x})$
  \Comment{Table~\ref{tab:classify}; total by its terminal row}
\State $M \gets \{$model terminal, output budget, tool loop, refusal,
  invalid tool, route capability$\}$
\State $T \gets \{$transport, incomplete SSE, provider envelope$\}$
\State $J \gets \{$invalid JSON, envelope violation, empty final,
  incomplete model response$\}$
\State $H \gets \{$permanent provider, evaluator risk, unclassified$\}$
\If{$c = \textsc{UnsupportedCapability}$}
  \State $\delta \gets \textsc{Coverage}$
  \Comment{declared unsupported work is never converted to zero}
\ElsIf{$c \in M$}
  \State $\delta \gets \textsc{Score}$
  \Comment{measured system behaviour; never retried}
\ElsIf{$c \in T$}
  \State $\delta \gets \textsc{Recover}$; reissue the exact request until the
    first provider response or the cap
    $c_{\text{adj}} = 4$ recovery attempts, as fixed in the signed recovery registry
  \State the first-pass failure \emph{remains} in the primary score; any
    recovered value is reported separately as transport-corrected and never
    replaces it
\ElsIf{$c \in J$}
  \State $\delta \gets \textsc{Retry}$; exactly one exact-request retry
  \State the first-pass failure \emph{remains} in the primary score; the retry
    outcome is retained as qualitative reliability evidence only and enters no
    score
\ElsIf{$c \in H$}
  \State $\delta \gets \textsc{Hold}$
  \Comment{repair/reseal or gold-blind manual review; no task substitution}
\Else
  \State $\delta \gets \textsc{Hold}$
  \Comment{fail closed: an unmodelled class is never silently scored}
\EndIf
\State \textbf{assert} a contract-valid but incorrect answer is never eligible
  for any branch above
\State \Return $\delta$
\end{algorithmic}
\end{algorithm}

\begin{table}[!tp]
\caption{$\textsc{Classify}$, the decision table of
Algorithm~\ref{alg:adjudicate}. Evaluated top to bottom; first match wins.
\emph{Envelope} is the sanitized provider-envelope summary carried in
$\hat{x}$; ``--'' means the row does not test it. Status codes are the five the
result schema admits, abbreviated: \textsc{err} \texttt{error}, \textsc{tmo}
\texttt{timeout}, \textsc{inv} \texttt{invalid\_output}, \textsc{uns}
\texttt{unsupported}. Coverage codes are \textsc{att}
\texttt{supported\_attempted} and \textsc{uns} \texttt{unsupported}. No column of this table is
derivable from a gold answer, a grader output, or a score, which is the
property Section~\ref{sec:adjudication} claims. A result with coverage state
\texttt{not\_run} never reaches this table: it was not attempted, so there is
nothing to adjudicate.}
\label{tab:classify}
\centering
\footnotesize
\setlength{\tabcolsep}{3.5pt}
\begin{tabular}{@{}rll>{\raggedright\arraybackslash}p{4.9cm}>{\raggedright\arraybackslash}p{2.05cm}l@{}}
\toprule
\# & Status & Coverage & Envelope predicate & Causal class & $\delta$ \\
\midrule
 1 & \textsc{uns} & \textsc{uns} & -- & unsupported capability & \textsc{Coverage} \\
 2 & \textsc{err} & \textsc{att} & permanent account, authorization or billing rejection & permanent provider & \textsc{Hold} \\
 3 & \textsc{err} & \textsc{att} & evaluator-flagged parameter incompatibility & evaluator risk & \textsc{Hold} \\
 4 & \textsc{err} & \textsc{att} & provider states a modality, tool or context limit of the route & route capability & \textsc{Score} \\
 5 & \textsc{err} & \textsc{att} & terminal refusal or safety stop & refusal & \textsc{Score} \\
 6 & \textsc{inv} & \textsc{att} & tool call names a tool outside the declared catalog, or arguments do not parse & invalid tool & \textsc{Score} \\
 7 & any & \textsc{att} & evaluator tool budget reached with no final-answer turn & tool loop & \textsc{Score} \\
 8 & any & \textsc{att} & \texttt{finish\_reason=length} at the frozen completion cap & output budget & \textsc{Score} \\
 9 & \textsc{err} & \textsc{att} & envelope carries a terminal model stop with no content and no error object & model terminal & \textsc{Score} \\
10 & \textsc{err}, \textsc{tmo} & \textsc{att} & no provider envelope was received & transport & \textsc{Recover} \\
11 & \textsc{err}, \textsc{tmo} & \textsc{att} & stream opened but no terminal event arrived & incomplete SSE & \textsc{Recover} \\
12 & \textsc{err} & \textsc{att} & envelope present and carries a provider-side error object & provider envelope & \textsc{Recover} \\
13 & \textsc{inv} & \textsc{att} & final message does not parse as JSON & invalid JSON & \textsc{Retry} \\
14 & \textsc{inv} & \textsc{att} & parses as JSON but fails the declared schema & envelope violation & \textsc{Retry} \\
15 & \textsc{inv} & \textsc{att} & final message is empty & empty final & \textsc{Retry} \\
16 & \textsc{inv} & \textsc{att} & terminal event present but no final-answer turn was produced & incomplete model response & \textsc{Retry} \\
17 & any & any & \emph{otherwise} & unclassified & \textsc{Hold} \\
\bottomrule
\end{tabular}
\end{table}

\subsection{Comparability: a run may only be extended by an identical run}
\label{sec:comparability}

A run may be extended only by work that would have produced the same run.
Algorithm~\ref{alg:resume} canonicalizes the run plan, which enumerates the task
set, the rendered requests, the system configuration, the corpus manifest and
the material harness contract; hashes it; and admits an append only when that
hash equals the one persisted at the run's checkpoint. Any difference forces a
new run rather than a partial merge, including a difference the evaluator
believes immaterial, because the predicate is equality of the hash and not a
judgement about what mattered. There is no per-task substitution: a single
re-run task cannot be spliced into a completed run. That is what licenses the
claim in Section~\ref{sec:costlat} that every published row is one complete
signed run under one route.

\begin{algorithm}[t]
\caption{Fail-closed comparability. An append is admitted only when the
candidate canonical run-plan hash exactly equals the persisted checkpoint.}
\label{alg:resume}
\begin{algorithmic}[1]
\Require existing run $R$ with immutable checkpoint $C_R$, candidate work $W$
\State $J(W) \gets$ the run-plan object enumerated in
  Appendix~\ref{app:manifest}
\State $\kappa(W) \gets \operatorname{SHA256}(\operatorname{JCS}(J(W)))$
  \Comment{UTF-8 JSON, sorted keys, compact separators, no ASCII coercion}
\State recompute and verify $\kappa(R)$ from $C_R.\texttt{plan}$
\If{$\kappa(W) = C_R.\texttt{run\_plan\_sha256} = \kappa(R)$}
  \State verify every retained task/request hash and reject duplicates or task
    identifiers outside $J(W).\texttt{task\_ids}$
  \State admit $W$ into $R$
\Else
  \State \textbf{reject}; $W$ must begin a new run
  \Comment{no partial merge, no per-task substitution}
\EndIf
\end{algorithmic}
\end{algorithm}

\subsection{Scoring}
\label{sec:scoring}

Task score $T_i$ is the predeclared-weight mean of its assertion credits,
scaled to 100. Suite score $S_k$ is the arithmetic mean of its locked task
scores, and the composite is the equal-weight mean over suites, which stops the
22-task suites dominating the smaller S1 and S7. The primary measure is the
first-pass, reliability-inclusive, equal-suite \ibib-7 Full score.

Reliability inclusion is the part that matters. Every attempted result is
passed once to the same deterministic assertion grader. A failure with no
parseable response earns zero because all declared assertions fail; a parseable
but contract-invalid response earns credit for the work products it actually
supplies, and its causal class adds no points and subtracts none. The only
fixed value is the mechanically implied zero for an absent response, never a
class-dependent penalty. An unsupported task is not attempted and is never
converted to a zero. Appendix~\ref{app:algorithms} gives the formulas, the weight-authoring rule per
suite, and the token-efficiency diagnostic.

\subsection{Sensitivity: what a score difference survives}
\label{sec:sensitivity}

Algorithm~\ref{alg:bootstrap} resamples tasks within suites, 20{,}000 draws per
pair under a pair-specific seed, and computes each interval once per unordered
pair in canonical order. Two cautions belong with the widths. Equal-suite
weighting means a suite's resampling noise enters in inverse proportion to its
task count, so widths are driven disproportionately by the smallest suites, S1
and S7. And a percentile bootstrap of a bounded mean has known coverage error
in small strata, which is exactly the regime the two thinnest cuts sit in. We
therefore recomputed both with the bias-corrected and accelerated variant over
retained per-task scores. The T1$\vert$T2 cut survives, $[0.11,6.80]$ becoming
$[0.32,7.04]$; the T2$\vert$T3 cut does not, $[0.05,7.62]$ becoming
$[-0.12,7.45]$ and crossing zero. An independent auditability limit is larger
than both thin lower bounds. R17 retained an aggregate count of two successful
Stage~0 reissues but not their task or suite identities. Under equal-suite
weighting, their placement-conditioned maximum leverage is 1.30 points if both
were in a 22-task suite and 2.86 points if both were in the 10-task suite; with
no attribution, 2.86 is the conservative cohort-level bound. The source values
and arithmetic are released as \texttt{paper/data/stage0\_sensitivity.csv}.
Because the other ten adapters did not retain zero-inclusive Stage~0 counts,
the relative bias is unknown. We therefore do not interpret either 0.11 or
0.05 as evidence for a distinct top cut, irrespective of bootstrap variant.
The regime is worth naming: under equal-suite
weighting S1 contributes roughly $2.2\times$ the variance per task that S2 does,
it is one of the four suites Section~\ref{sec:saturation} reports as saturated,
and it carries four systems tied at exactly 95.50, so its resampled mean is both
small-$n$ and strongly non-normal. The realized parameters and bounds are
released as \texttt{paper/data/bca\_t1\_t2\_terra\_qwen38.json} and
\texttt{paper/data/bca\_t2\_t3\_qwen38\_luna.json}.

\begin{algorithm}[t]
\caption{Paired suite-stratified task-set bootstrap for a score difference. Used
only to describe sensitivity to the selected task set, never as a confidence
interval for stochastic run-to-run variance.}
\label{alg:bootstrap}
\begin{algorithmic}[1]
\Require per-task scores $T^{A}_i, T^{B}_i$ for systems $A, B$ over identical
  task keys; suites $k = 1 \ldots K$ with sizes $n_k$; $B_{\text{iter}} = 20{,}000$;
  seed $= \operatorname{uint32}(\operatorname{SHA256}(m_A\|\texttt{|}\|m_B)_{1:8})$
\Ensure interval for $\Delta = \text{\ibib-7}(A) - \text{\ibib-7}(B)$
\For{$b = 1$ to $B_{\text{iter}}$}
  \For{$k = 1$ to $K$}
    \State draw $n_k$ task keys from suite $k$ \emph{with replacement},
      preserving $n_k$
    \State the \emph{same} keys are used for $A$ and $B$ \Comment{pairing}
    \State $S_k^{A(b)}, S_k^{B(b)} \gets$ means over the resampled keys
  \EndFor
  \State $\Delta^{(b)} \gets \frac{1}{K}\sum_k S_k^{A(b)} - \frac{1}{K}\sum_k S_k^{B(b)}$
\EndFor
\State \Return the 95\% percentile interval
  $[Q_{0.025}(\{\Delta^{(b)}\}),Q_{0.975}(\{\Delta^{(b)}\})]$
\Statex \hspace{-1.2em}\parbox{0.97\linewidth}{\footnotesize The pair-specific
seed uses the first eight hexadecimal digits of SHA-256, which makes the draw
depend on the order the two identifiers are supplied in. An interval is
therefore computed exactly once per \emph{unordered} pair, in canonical order
(the two model identifiers sorted lexicographically), and negated with its
bounds swapped when displayed in the other direction. Table~\ref{tab:main}'s
leader interval $[-2.13,4.26]$ is the negation of the stored
$[-4.26,2.13]$ under canonical seed 224349676; it was not recomputed under a
second seed. The signed report records the realized canonical seed for every
pair.}
\end{algorithmic}
\end{algorithm}

\section{The reference instantiation and how it was run}
\label{sec:construction}

\subsection{Suites and profiles}
\label{sec:suites}

Seven suites hold 128 locked tasks
and 987 deterministic assertions over enterprise document, spreadsheet, chart,
tool and governed-database work. The suites are single-tab and multi-tab
spreadsheet reasoning (S1, S2), selectable-text and image-only document analysis (S3, S4), chart and
diagram interpretation (S5), tool calling in a stateful evaluator-owned sandbox
(S6), and governed database work against a ten-million-row SQLite instance with
a field dictionary of certified columns hidden among decoys (S7). Two profiles
are exposed: \ibib-6 Core covers all but S4 for 106 tasks, and \ibib-7 Full
adds S4 for 128. S4 is separable because OCR is often supplied by a component
outside the model; S5 is not, because chart interpretation is a capability of
the system rather than of a preprocessing component. Treating them together
would make S5 simultaneously a Core requirement and a droppable lane.
Every task is authored with a typed output contract, citation requirements,
deterministic gold assertions with numeric tolerances and a privacy
classification. Task and gold files stay physically separate, so gold cannot
enter a provider-visible request. Approximately 30\% of each suite is held as a
challenge reserve for saturation detection. Appendix~\ref{app:instrument} gives
suite-by-suite construction, the task model and the contamination controls.

\begin{definition}[Suite classification]
\label{def:sat}
Let the reference band $B$ be the six systems with the highest \ibib-7 Full
scores, and let $b_{(1)} \leq \cdots \leq b_{(6)}$ be their scores on the suite.
Define the median $\tilde{b} = (b_{(3)}+b_{(4)})/2$, the count
$n_{90} = |\{b \in B : b \geq 90\}|$, and the trimmed range
$\rho = b_{(5)} - b_{(2)}$. A suite is
\begin{itemize}[leftmargin=1.4em,itemsep=0pt,topsep=2pt]
  \item \textbf{saturated} if $\tilde{b} \geq 90$ and $n_{90} \geq 4$;
  \item otherwise \textbf{discriminative} if $\rho \geq 10$;
  \item otherwise \textbf{compressed below ceiling}: it separates nothing in the
    band and is not solved by it either.
\end{itemize}
\end{definition}

\subsection{Difficulty, admission, and the anti-ceiling policy}
\label{sec:anticeiling}

Difficulty must come from the work. Every task must be answerable from the
supplied evidence and permitted tools; ambiguous wording, illegible inputs and
adversarial grading are not valid sources of it. The realized difficulty labels
do not survive contact with the results. With 93.8\% of tasks labelled Expert
or Frontier while four of seven suites are saturated, and with the labels
effectively constant within a suite, they are non-predictive of measured
difficulty. Nothing in this paper's results rests on them, so we treat the
taxonomy as an unvalidated construct in v0.13.2 and make blind difficulty
piloting a freeze requirement (Section~\ref{sec:frozen}) rather than defend the
labels here. Appendix~\ref{app:instrument} gives the target profile, the admission gate, the
anti-ceiling policy and the band-size sensitivity of
Definition~\ref{def:sat}.

\subsection{Cohort and execution}
\label{sec:setup}

Table~\ref{tab:systems} lists the eleven evaluated systems. The cohort is
purposive rather than exhaustive: three GPT-5.6 endpoints at \texttt{xhigh},
two Anthropic operating points, three Qwen models spanning a sparse and two
dense configurations, Kimi K3, Muse Glimmer 30B, and the composed GLM system
of Section~\ref{sec:sut}. This is the cohort frozen before execution;
every open-weight system uses exactly one serving route, per
Property~\ref{prop:nonsub}.

Tasks are issued serially at concurrency one. Latency is evaluator-observed per
task and includes provider queueing and evaluator-owned work, so it is not
model-only inference latency. Cost is a measured run diagnostic rather than a
reconciled invoice. Sampling parameters were frozen no later than
2026-07-23~05:50:04~UTC, which is also when the first retained
complete-candidate run began, so the evidence bounds the freeze by the first
launch rather than attesting it independently; a dated freeze attestation is a
freeze requirement. Several determinants of behaviour are not observable on a
hosted route at all, including inference engine and version, guided-decoding
backend, KV-cache policy and hardware generation, and every one of them is
known to move accuracy for identical weights. That is why the protocol binds
and records what \emph{is} observable and defers attribution rather than
inferring it. Appendix~\ref{app:instrument} gives the cohort rationale, the observability
split and the cost accounting in full.

\begin{table}[t]
\caption{Evaluated systems. Column headers carry the tuple symbols of
Definition~\ref{def:sut}. $\mathcal{T}$, $\mathcal{C}$ and $H$ are identical for
every row by construction, and $\Omega$ is the contract of
Table~\ref{tab:contract} as \emph{observed} in binding rather than as
advertised. GPT-5.6 Sol, GPT-5.6 Terra, and GPT-5.6 Luna are distinct pinned model identifiers, printed
in the first column rather than treated as settings of one identifier. The GLM
row is a composed system in the sense of
Definition~\ref{def:sut}, with $\sigma$ mapping S4 and S5 to GLM-5V-Turbo and
every other suite to GLM-5.2.}
\label{tab:systems}
\centering
\footnotesize
\begin{tabular}{@{}lllp{3.9cm}l@{}}
\toprule
Model $m$ & Effort $e$ & Precision $p$ & Route $r$ & Vision path \\
\midrule
GPT-5.6 Sol   & xhigh & provider & OpenAI Responses, \texttt{store=false} & native \\
GPT-5.6 Terra & xhigh & provider & OpenAI Responses, \texttt{store=false} & native \\
GPT-5.6 Luna  & xhigh & provider & OpenAI Responses, \texttt{store=false} & native \\
Claude Fable 5  & high & provider & Anthropic Messages & native \\
Claude Sonnet 5 & high & provider & Anthropic Messages & native \\
Qwen3.5-122B-A10B & high  & BF16  & OpenRouter $\to$ Novita, no fallback      & native \\
Qwen3.6-27B       & high  & FP8   & OpenRouter $\to$ Provider P, no fallback & native \\
Qwen3.8-27B       & xhigh & BF16  & OpenRouter $\to$ AkashML, no fallback     & native \\
Kimi K3           & max   & MXFP4 & OpenRouter $\to$ Modal, no fallback       & native \\
GLM-5.2           & xhigh & FP8   & OpenRouter $\to$ Z.AI, no fallback        & routed to \\
\quad + GLM-5V-Turbo & high & FP8 & OpenRouter $\to$ Z.AI, no fallback        & \quad GLM-5V \\
Muse Glimmer 30B  & xhigh & BF16  & direct DeepInfra & native \\
\bottomrule
\end{tabular}
\end{table}

\section{Results}
\label{sec:results}

All results were generated after every run completed, the gold-blind
adjudication ledger cleared its publication holds, and the fail-closed
comparability audit passed for all eleven systems. Some completed runs
originated under the immediately preceding harness contract; for those, a
signed equivalence bridge attests agreement on the enumerated surface of
provider-visible task bytes, tool implementations, route configuration, budgets
and scoring semantics, and $H$ is identified up to that equivalence rather than
byte-identity.

\subsection{Composite results: nominal tiers and evidentiary groups}
\label{sec:tiers}

Table~\ref{tab:main} gives the primary scores. We report resolution groups rather than
ranks. The leader-minus-runner-up difference is 0.97 displayed points against a
task-set sensitivity interval of $[-2.13,4.26]$, which straddles zero, and the
same holds at positions seven and eight. We computed all 55 pairwise intervals
and formed tiers greedily from the point-estimate order: a candidate begins a
new tier only when its interval excludes zero against every member of the
current tier. The five labels printed in the table are the mechanical output of
that nominal rule and are retained as a protocol diagnostic, not as five
inferentially distinct strata.

That construction makes 55 simultaneous decisions at a nominal 95\% level with
no multiplicity adjustment, and we report it at that level rather than an
adjusted one. Forty-six of the 55 intervals exclude zero. The two thinnest in
the whole matrix are exactly the two cuts splitting the top nine: the
T1$\vert$T2 cut, GPT-5.6 Terra minus Qwen3.8-27B at $[0.11,6.80]$, a margin of
1.6\% of the interval width, and the T2$\vert$T3 cut, Qwen3.8-27B minus GPT-5.6
Luna at $[0.05,7.62]$, a margin of 0.7\%. The second does not survive the BCa
recomputation of Section~\ref{sec:sensitivity}. Neither survives any multiplicity adjustment;
Bonferroni at 55 comparisons requires roughly a 99.9\% interval. The greedy
rule then merges T1, T2 and T3 into one group spanning 88.34 to 68.27. The two
lower boundaries are not close calls, clearing zero by 66\% and 39\% of their
widths. The Stage~0 bound in Section~\ref{sec:sensitivity} independently makes
both top cuts uninterpretable. Accordingly, the primary evidentiary reading is
that the cohort separates into $\{$top nine$\}$, Qwen3.5-122B-A10B, and
GLM-5.2 + GLM-5V-Turbo, and that the three-way split \emph{inside} the top nine
is diagnostic only. In particular, we do not claim Qwen3.8-27B as an
evidentially distinct T2 group.

\begin{table}[t]
\caption{Primary results, ordered by \ibib-7 Full. \emph{Valid} is the number of
the 128 tasks that returned a scoreable response; it is distinct from
\emph{coverage} in Table~\ref{tab:ops}, which is the number of capability lanes
a system supports and is complete for every system in this cohort. \emph{Gap} is the
magnitude of the difference to the row above; every gap is stated as
higher-minus-lower so that no sign convention is needed. All 55 pairwise
comparisons use Algorithm~\ref{alg:bootstrap}. The four boundary intervals,
top to bottom, are $[0.11,6.80]$, $[0.05,7.62]$, $[8.48,21.35]$ and
$[5.37,19.17]$; Section~\ref{sec:tiers} reads them and states the multiplicity
and BCa sensitivity. The displayed nominal groups retain the preregistered
percentile rule only as a reproducible diagnostic; the paper's evidentiary
interpretation merges T1--T3, also because R17's unattributable Stage~0 exposure
has 1.30--2.86 points of placement-conditioned maximum leverage. Within-group
separation counts make ``group'' concrete:
0 of 3 within-T1 pairs exclude zero, and 4 of 10 within-T3 pairs do
(GPT-5.6 Luna--Kimi K3, GPT-5.6 Luna--Qwen3.6-27B,
Claude Sonnet~5--Qwen3.6-27B, Muse Glimmer 30B--Qwen3.6-27B). The complete
matrix and realized pair seeds are released as
\texttt{paper/data/pairwise\_intervals.csv}.}
\label{tab:main}
\centering
\small
\begin{threeparttable}
\begin{tabular}{@{}llrrrl@{}}
\toprule
Nominal & System & \ibib-7 Full & \ibib-6 Core & Valid & Gap to row above \\
\midrule
\multirow{3}{*}{T1} & GPT-5.6 Sol    & 88.34 & 87.25 & 127/128 & \no \\
                    & Claude Fable 5 & 87.37 & 86.11 & 128/128 & 0.97 \\
                    & GPT-5.6 Terra & 85.95 & 83.98 & 127/128 & 1.42 \\
\midrule
T2 & Qwen3.8-27B & 82.54 & 80.30 & 126/128 & 3.41 \\
\midrule
\multirow{5}{*}{T3} & GPT-5.6 Luna     & 78.65 & 75.66 & 117/128 & 3.89 \\
                    & Claude Sonnet 5  & 75.40 & 72.23 & 126/128 & 3.25 \\
                    & Muse Glimmer 30B & 73.66 & 71.33 & 122/128 & 1.74 \\
                    & Kimi K3          & 72.99 & 69.90 & 108/128 & 0.67 \\
                    & Qwen3.6-27B\tnote{$\ddagger$} & 68.27 & 65.39 & 101/128 & 4.72 \\
\midrule
T4 & Qwen3.5-122B-A10B & 53.32 & 50.95 & 102/128 & 14.94 \\
\midrule
T5\tnote{$\dagger$} & GLM-5.2 + GLM-5V-Turbo & 41.10 & 38.11 & 86/128 & 12.22 \\
\bottomrule
\end{tabular}
\begin{tablenotes}[flushleft]\footnotesize
\item Transport recovery under Algorithm~\ref{alg:adjudicate} changed the score
for two systems: Muse Glimmer 73.66$\to$74.88 and Kimi K3 72.99$\to$76.93.
GLM's single transport failure did not change its score, which remains 41.10.
No other primary score in this table is transport-corrected, Qwen3.8-27B's
82.54 included; see Section~\ref{sec:reliability}. Corrected scores are
reported separately and never replace the primary score.
\item[$\dagger$] GLM's S6 and S7 scores are measured final-answer contract
failures of this served route and are scored as such; they are not attributed
to the raw GLM weights. See Section~\ref{sec:glm}. Every composite in this
table reproduces to 0.01 from the suite scores of Table~\ref{tab:suitescores}.
\emph{Gap} is a difference of the displayed two-decimal values; the
full-precision ledger gives the leading pair as 0.96 and GLM's \ibib-6 Core as
38.11, and that ledger, not the displayed values, is the authoritative rounding
path. Qwen3.6-27B's composite is 68.265866 and is therefore printed as 68.27
here; the released aggregate extracts under \texttt{paper/data/} carry the
two-decimal truncation 68.26 for that row, so a reader comparing the two will
find a 0.01 discrepancy that resolves in favour of this table.
\item[$\ddagger$] Qwen3.6-27B's route is withheld (Provider~P,
Section~\ref{sec:ethics}), so this row is under-identified under
Definition~\ref{def:sut}, which makes the route a component of the system under
test, and it is not reproducible in principle by a reader. See
Section~\ref{sec:naming}.
\end{tablenotes}
\end{threeparttable}
\end{table}

\subsection{Where discrimination actually lives}
\label{sec:suiteresults}

Table~\ref{tab:suitescores} is the paper's most consequential table and it
answers RQ3. Definition~\ref{def:sat} reports four of seven suites saturated:
S6 (median 96.9, $n_{90}=5$), S4 (95.5, 6), S1 (93.3, 4) and S3 (92.1, 6). Two
discriminate, S7 ($\rho = 31.4$) and S2 ($\rho = 12.5$). One, S5, is compressed below ceiling: no band system reaches 90 and $\rho$ is
only 6.0. That third case is worth naming separately, because a single spread
statistic hides it. A suite that neither separates frontier systems nor has been
solved by them is not evidence of a hard construct and not evidence of a
saturated one; it is a suite whose assertions are probably not measuring what
its difficulty labels claim.

Two consequences follow, both against our own interest. S4, the image-only OCR
lane that was the most expensive suite in the bank to build, is the least
discriminative suite it contains. And the two discriminating suites carry the
composite: the headline range from 41.10 to 88.34 is produced by governed
database work, by multi-tab joins, and by weaker systems failing S6 outright,
not by the multimodal document design that motivates four of the seven suites.
The lower end carries a qualifier: GLM's 41.10 is the score of a route whose S6
and S7 failures are strict-JSON contract failures under \emph{our} published
contract, so the range is contract-dependent at its bottom in a way it is not
at its top (Section~\ref{sec:glm}).

\begin{table}[t]
\caption{Suite-level scores. Systems are ordered by \ibib-7 Full. The three
right-hand columns are the saturation evidence and are computed over the
\emph{top six} systems only, $\{$GPT-5.6 Sol, Claude Fable~5, GPT-5.6 Terra,
Qwen3.8-27B, GPT-5.6 Luna, Claude Sonnet~5$\}$, so weak-system collapse cannot
inflate them. Compact column headers are used only to fit the table; they map,
in order, to the official system names in Table~\ref{tab:systems}. The
untrimmed range is not reported because over six points one system sets it:
S6's is 27.0 against a trimmed 10.0, the difference being Claude Sonnet~5
alone. \emph{Med.}
is the band median, $n_{90}$ the number of band systems scoring at least 90,
and $\rho$ the trimmed range $b_{(5)} - b_{(2)}$; Definition~\ref{def:sat}
classifies from these three. Rows are grouped by class and ordered by $\rho$
within each group. Derived
values shown here are rounded from the full-precision aggregate ledger; derived
composites are not recomputed from already-rounded table cells.}
\label{tab:suitescores}
\centering
\scriptsize
\setlength{\tabcolsep}{1.7pt}
\begin{threeparttable}
\begin{tabular}{@{}lrrrrrrrrrrr@{\hspace{9pt}}rrrl@{}}
\toprule
& \multicolumn{6}{c}{top six} & \multicolumn{5}{c}{} & \multicolumn{4}{c}{top six only} \\
\cmidrule(lr){2-7}\cmidrule(l){13-16}
Suite & Sol & Fable & Terra & Q3.8 & Luna & Son.5 & Muse & Kimi & Q3.6 & Q3.5 & GLM & Med. & $n_{90}$ & $\rho$ & Class \\
\midrule
S7 governed DB    & 53.66 & 67.47 & 53.16 & 51.59 &  7.29 & 22.22 & 52.26 & 29.66 & 12.10 &  5.29 &  0.00 & 52.4 & 0 & 31.4 & discriminative \\
S2 multi-tab      & 93.75 & 91.48 & 80.68 & 78.98 & 79.55 & 76.70 & 61.51 & 76.70 & 70.60 & 49.43 & 42.05 & 80.1 & 2 & 12.5 & discriminative \\
\midrule
S5 charts         & 85.91 & 86.82 & 82.73 & 80.00 & 79.20 & 79.89 & 80.80 & 85.11 & 80.11 & 42.50 & 54.77 & 81.4 & 0 &  6.0 & compressed$^{a}$ \\
\midrule
S6 tools          &100.00 & 93.76 &100.00 & 90.00 &100.00 & 72.96 & 87.22 & 40.00 & 68.33 & 88.89 &  5.00 & 96.9 & 5 & 10.0 & \textbf{saturated}$^{b}$ \\
S1 single-tab     & 95.50 & 86.50 & 95.50 & 88.75 & 95.50 & 91.00 & 55.25 & 95.50 & 70.30 & 58.05 & 62.55 & 93.3 & 4 &  6.8 & \textbf{saturated}$^{c}$ \\
S3 select.\ PDF   & 94.66 & 90.64 & 91.82 & 92.46 & 92.39 & 90.64 & 90.95 & 92.42 & 90.91 & 61.52 & 64.32 & 92.1 & 6 &  1.8 & \textbf{saturated} \\
S4 image OCR      & 94.89 & 94.94 & 97.73 & 96.02 & 96.59 & 94.37 & 87.65 & 91.53 & 85.50 & 67.53 & 59.01 & 95.5 & 6 &  1.7 & \textbf{saturated} \\
\bottomrule
\end{tabular}
\begin{tablenotes}[flushleft]\scriptsize
\item $^{a}$ S5 falls through both tests of
Definition~\ref{def:sat}: no band system reaches 90, and $\rho = 6.0$. It is the
one suite currently buying the bank neither discrimination nor a demonstration
of capability.
\item $^{b}$ Five of six band systems at or above 90, three at exactly
100.00. S6's untrimmed range is 27.0 against $\rho = 10.0$; the difference is
Claude Sonnet~5 alone.
\item $^{c}$ Four systems tied at exactly 95.50 on a 10-task suite,
three of them in the top six. One task is worth 10 points on S1, so the two
printed decimals are an artifact of the composite arithmetic, not a claim of
resolution.
\end{tablenotes}
\end{threeparttable}
\end{table}

\begin{figure}[t]
\centering
\includegraphics[width=.95\textwidth]{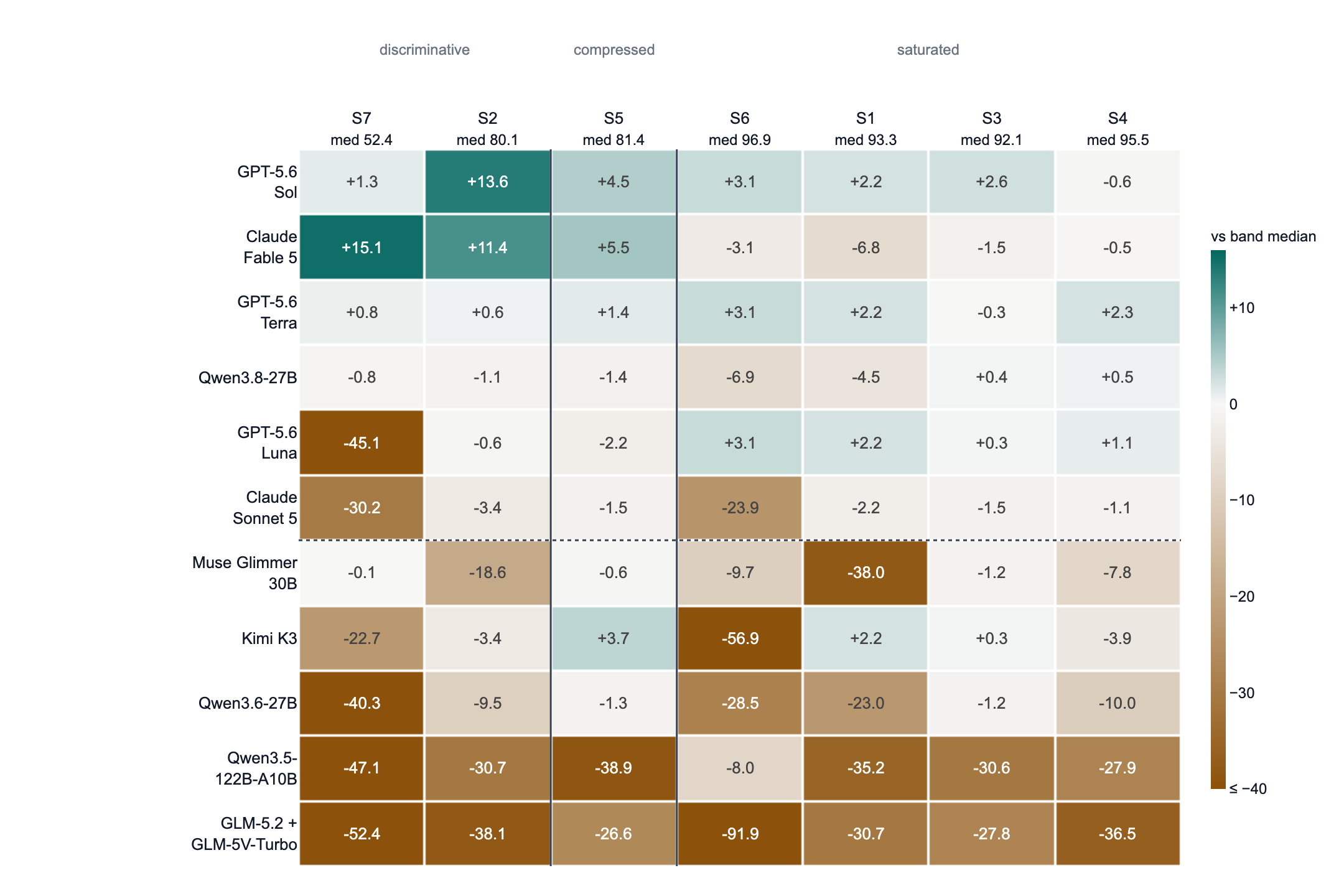}
\caption{Suite profiles as \emph{deviation from the top-six band median},
which is the quantity Definition~\ref{def:sat} classifies on and the one
Table~\ref{tab:suitescores} cannot show at a glance; each column header carries
its band median, so any cell's raw score is the header plus the printed
deviation. Systems are ordered by \ibib-7 Full and columns follow
Table~\ref{tab:suitescores}'s ordering rule exactly. Vertical rules separate the
discriminative suites (S7/S2), the compressed-below-ceiling suite (S5), and the
four saturated suites; the dotted horizontal rule marks the bottom of the
six-system band the medians are computed over. The four right-hand columns are
near-white across the band, which is what ``nearly uniform at the frontier''
means, while S7 and S6 carry deviations of tens of points for systems that are
otherwise close. Colour saturates at $\pm 40$ so that one cell does not set
the scale for all 77; six cells exceed it, and every cell prints its own exact
deviation regardless.}
\label{fig:heatmap}
\end{figure}

\subsection{Reliability and failure evidence}
\label{sec:reliability}

Table~\ref{tab:failures} gives the failure taxonomy as a system $\times$ class
matrix, 138 first-pass failures in total. Three columns carry most of the
interpretation. Tool-loop exhaustion is a planning failure against a fixed call
budget and is concentrated in GPT-5.6 Luna, which accounts for eleven of the
33. The three contract classes are integration failures and are concentrated in
Kimi K3, Qwen3.6-27B and the composed GLM system. Budget exhaustion is model
behaviour and is concentrated in that same GLM system. An operator reading
only the composite would see none of this.

Eleven objectively classified transport failures received exact-request
recovery, and 82 malformed responses received the single permitted supplemental
retry as qualitative evidence: 41 invalid JSON-contract responses, 23 envelope
violations and 18 empty finals. In every case the first-pass failure stays in
the primary score and any value recovered afterwards is reported separately.

One asymmetry is on the record rather than smoothed over. Stage~0, the
request-level reissue that runs inside the harness before a result record
exists, leaves no task-level mark when it succeeds within its cap. The R17
adapter retained only a run-level count of two successful reissues; the failed
pre-envelope attempts, task identifiers and suite identifiers were not
persisted, so exact attribution is not recoverable. R17's recorded transport
failure count is therefore zero, but it is not comparable to the other ten
systems' counts: their historical adapters did not persist zero-inclusive
Stage~0 counters at all. Qwen3.8-27B's published 82.54 is \emph{not}
transport-corrected, while Muse Glimmer's three exhausted transport faults and
Kimi K3's seven remain in their primary scores.

The two R17 reissues have 1.30--2.86 points of placement-conditioned maximum
leverage and a fully conservative zero-first-attempt lower envelope of 79.68,
as detailed in Section~\ref{sec:sensitivity}. This is larger than the 0.11 and
0.05 lower bounds defining the nominal top cuts, so those cuts are not treated
as evidentiary. Applying the separate Stage~1 transport recovery uniformly to
the systems for which it exists moves Kimi K3 from eighth to sixth by point
score, ahead of Claude Sonnet~5 and Muse Glimmer 30B. We report that reordering
as a supplemental reliability diagnostic but decline to replace the primary
cohort ordering with it: successful Stage~0 exposure cannot be corrected
uniformly across systems. Frozen-v1 requires a zero-inclusive request-attempt
count on every result record.

\begin{table}[t]
\caption{Failure taxonomy as a system $\times$ class matrix. A column is a
failure class, a row a system's reliability signature. \emph{Retry
eligibility} follows Algorithm~\ref{alg:adjudicate}: transport failures are
recovered and reported separately, the three contract classes receive one
supplemental retry, and budget exhaustion and tool-loop exhaustion are scored as
model behaviour and never retried. The matrix omits Stage~0 counts because ten
historical adapters did not persist them; R17's aggregate count of two and its
non-attributability are recorded in \texttt{paper/data/run\_dispositions.csv}
(Section~\ref{sec:reliability}).}
\label{tab:failures}
\centering
\small
\setlength{\tabcolsep}{4pt}
\begin{threeparttable}
\begin{tabular}{@{}lrrrrrrr@{}}
\toprule
 & \multicolumn{1}{c}{Transp.} & \multicolumn{3}{c}{contract classes (retried once)} & \multicolumn{2}{c}{scored, never retried} & \\
\cmidrule(lr){2-2}\cmidrule(lr){3-5}\cmidrule(lr){6-7}
System & failure & JSON & envelope & empty & tool loop & budget & Total \\
\midrule
GPT-5.6 Sol       & 0 &  0 &  0 &  0 &  1 & 0 & 1 \\
Claude Fable 5    & 0 &  0 &  0 &  0 &  0 & 0 & 0 \\
GPT-5.6 Terra     & 0 &  0 &  0 &  0 &  1 & 0 & 1 \\
Qwen3.8-27B       & 0 &  0 &  0 &  0 &  0 & 2 & 2 \\
GPT-5.6 Luna      & 0 &  0 &  0 &  0 & 11 & 0 & 11 \\
Claude Sonnet 5   & 0 &  0 &  0 &  0 &  2 & 0 & 2 \\
Muse Glimmer 30B  & 3 &  2 &  0 &  0 &  0 & 1 & 6 \\
Kimi K3           & 7 & 13 &  0 &  0 &  0 & 0 & 20 \\
Qwen3.6-27B       & 0 &  0 & 16 & 10 &  0 & 1 & 27 \\
Qwen3.5-122B-A10B & 0 &  2 &  7 &  8 &  9 & 0 & 26 \\
GLM-5.2 + GLM-5V-Turbo & 1 & 24 &  0 &  0 &  9 & 8 & 42 \\
\midrule
\textbf{Total}    & \textbf{11} & \textbf{41} & \textbf{23} & \textbf{18} & \textbf{33} & \textbf{12} & \textbf{138} \\
\bottomrule
\end{tabular}
\end{threeparttable}
\end{table}

\subsection{The GLM tool lanes are route-level diagnostics}
\label{sec:glm}

GLM-5.2 + GLM-5V-Turbo recorded 5.00 on S6 and 0.00 on S7. We audited every
provider envelope without consulting gold answers or task scores. All 24
responses classified as JSON-contract failures carry a parser or format
signature rather than a reasoning one: 23 are a syntactically valid JSON object
wrapped in prose, and one contains malformed embedded JSON. Within S6, 17 of 18
tasks have the first signature and one returns raw contract-valid JSON. Within
S7, two have it, nine exhaust the preregistered thirteen-turn budget while
continuing to call tools, and one stream ends in an objective transport
interruption. The dated audit is released as
\texttt{paper/data/glm\_forensic\_review.csv}.

Those observations separate three phenomena the aggregate would conflate. The 24
strict-JSON failures are measured final-answer contract failures of this served
system; the nine loop exhaustions are agent behaviour under a fixed budget; the
interrupted stream is provider transport. None is reclassified as a wrong
substantive answer, and all stay in the score.

A third reading of the 23 prose-wrapped responses is available and we do not
think it can be dismissed: that the strictness is the evaluator's rather than the
route's. A production integration built around a lenient extractor would recover
most of them, and on that reading part of GLM's 41.10 is an artifact of where we
set the parser rather than a property of the served system. Our contract
(Table~\ref{tab:contract}) requires a bare schema-conformant final message and
was published before any run, so applying it is not post-hoc; but the choice of a
strict contract over a lenient one is ours. We did not run the counterfactual,
because rescoring the 23 under a permissive extractor would mean re-entering the
sealed scoring path after seeing scores, which Section~\ref{sec:adjudication}
forbids. So we neither report a leniently-extracted score nor claim one would be
low. Under \emph{this} contract the responses are failures, and a reader whose
own integration is more permissive should read GLM's S6 and S7 as a lower bound
(Section~\ref{sec:failuremeaning}).

\subsection{The binding gate has a named gap, and it is a finding about Algorithm~\ref{alg:bind}}
\label{sec:probegap}

Only an admission that later fails tests what a \textsc{Bound} disposition is
worth, and GLM is the one such case this cohort contains. The gate missed. This
route passed all ten predicates of Table~\ref{tab:probes} and then failed
strict-JSON adherence in production, so the gate is incomplete rather than
merely unlucky. Probe~3 verifies that an \emph{ordinary} turn returns a
schema-conformant envelope. It does not verify that the final message body
parses as bare JSON after the model has emitted extended reasoning. The missing
predicate is named at the foot of Table~\ref{tab:probes}. We did not run it
here and do not report it as though we had.

Read with the unreachable coverage branch of Section~\ref{sec:binding}, the two
gaps point the same way. A gate is only as good as the predicates it contains,
and this one has none that isolates the single separable lane and none that
tests structured output under the condition production actually imposes. Both
are cheap to add and neither was added in time for this cohort. That is the
most transferable lesson here for anyone implementing Algorithm~\ref{alg:bind}:
enumerate the predicate-to-lane mapping first, then check that every
disposition the algorithm can return is reachable from it.

\subsection{Cost, latency, efficiency, and run disposition}
\label{sec:costlat}

Among systems with complete cost telemetry, Muse Glimmer 30B recorded the
lowest full-run inference cost at \$2.43 and GPT-5.6 Terra the lowest median
task latency at 18.3 seconds. Token efficiency, in fully-correct-task
equivalents per million completion tokens, ranges from 171.81 to 22.46. The
measured score-cost Pareto frontier contains Muse Glimmer 30B, GPT-5.6 Luna,
Qwen3.8-27B, GPT-5.6 Terra and GPT-5.6 Sol; the score-latency frontier contains
GPT-5.6 Terra, Claude Fable~5 and GPT-5.6 Sol. Reasoning share of completion
tokens ranges from 53.5\% to 96.9\% and is not separately reported by Muse
Glimmer 30B; because providers tokenize and account for hidden reasoning
differently, that measure qualifies the ranking rather than producing a second
one. Both frontiers describe these endpoints in this run under this locked task mix.
They are not universal efficiency claims, and they establish no economic value,
analyst replacement or procurement threshold. Of the retained runs, every published row is
a complete signed run under one route, and no result is spliced from more than
one. Appendix~\ref{app:ops} gives the operating-characteristics table, the
efficiency frontiers and the full run-disposition ledger.

\section{Does the protocol change any conclusion?}
\label{sec:ablations}
\label{sec:serving}

Each subsection recomputes the cohort under a convention the protocol rejects.
None requires new inference: all are recomputations over retained
per-assertion results, or, in Ablation~A's first half, a reconciliation of two
complete retired route runs with later gold-blind binding receipts.

\subsection{Ablation A: route binding versus evaluate-by-model-identifier (RQ1)}
\label{sec:ablA}

RQ1 has two halves and the cohort answers them with different strength.

\paragraph{Executability.} R11 and R12 were complete 128-task calibration runs,
not routes rejected before locked release. They scored 26.12 on DeepInfra FP8
and 62.25 on CoreWeave FP8. The later finalized binding gate found DeepInfra
failed predicate~4 and CoreWeave failed predicate~7; Provider~P then passed the
full gate prospectively and scored 68.27 in a fresh whole-route run. The
complete predicate set postdates R12, so the first two capability findings are
retrospective and cannot support a claim of preregistered route exclusion.
Table~\ref{tab:ablroute} carries all three scores and the chronology.

The score deltas to Provider~P, $+42.14$ and $+6.01$, are descriptive paired
route contrasts over identical locked task identifiers and weights. Each route
was run once and no run-to-run variance interval exists, so we use neither
delta for an inferential tier claim. What survives is a reporting result: a
model-identifier benchmark cannot express that the first two complete scores
came from routes later shown unable to satisfy the finalized envelope. A route
that cannot serve thirteen images is not a bad route, only one that cannot
serve \emph{this} contract.

\paragraph{Profile and nominal tier diagnostic.} Qwen3.8-27B scored 77.38 on the fair-v5 RunPod arm
and 82.54 on the fair-v6 AkashML arm over identical task keys and weights, 5.16
equal-suite points apart, with a 95\% paired task-set interval of $[0.11,10.60]$
under canonical pair seed 4046266586. Neither arm is transport-corrected. We
report that as an \emph{endpoint} effect rather than a route effect.
Table~\ref{tab:armdiff} states the tuple difference component by component. The
arms differ in access mode, harness generation and the serving tool-call parser
as well as in $r$, so the 5.16
points are a joint effect of that column. Section~\ref{sec:expvariance} names
the study that would separate them.

Substituting the RunPod arm and recomputing all pairwise intervals keeps the arm
in nominal T2: its intervals against GPT-5.6 Luna, Claude Sonnet~5, Muse Glimmer 30B and
Kimi K3 all straddle zero, and it stays separated from every T1 system. What
moves is the diagnostic ladder around it. In the published cohort Qwen3.8-27B is the sole
occupant of nominal T2 and Luna heads T3; under substitution the arm falls below Luna, T2
and T3 merge into one six-member group spanning 78.65 to 68.27, and the nominal five-tier
ladder becomes a four-tier one. Endpoint choice does not move this arm across a
boundary, then, but the boundary it would have crossed is not robust to the
substitution in the first place.

That is the third independent reason this paper gives to distrust the T1--T3
split. Section~\ref{sec:tiers} shows that neither top cut survives multiplicity
adjustment across 55 comparisons; Section~\ref{sec:sensitivity} shows the
T2$\vert$T3 cut crossing zero under BCa; and the substitution here dissolves the
same cut a third way. The two boundaries below are untouched by all three. The
substituted ladder and all ten directed intervals against the arm are released as
\texttt{paper/data/route\_arm\_tiers.csv} and
\texttt{paper/data/route\_arm\_tiers\_pairs.csv}.

\subsection{Ablation B: reliability-inclusive versus accuracy-only (RQ2)}

The ordering changes materially: GPT-5.6 Luna becomes the point leader, while
Kimi K3 and Qwen3.6-27B move above systems with much higher first-pass
reliability. This is the answer to RQ2. Conditional accuracy describes
correctness after integration failure has been removed; it is not a substitute
for a deployment score.

\subsection{Coverage reporting versus zero-fill}
\label{sec:ablC}

The other three recompute over observations the cohort produced. There are
none here: every evaluated system has complete lane coverage and the
\textsc{PartialCoverage} branch never fired. Constructing the case by hand,
zero-filling two unsupported lanes for one system prints 57.40 and would place
it tenth, but that number fabricates two measured failures from two unmeasured
lanes; the difference from ``withheld'' is categorical, not 57.40 points. The
argument is sound and the rule is worth having. What it is not is evidence. The
coverage machinery has no empirical support anywhere in this paper: not its
disposition in Algorithm~\ref{alg:bind}, not its state in the result schema, not
its column in Table~\ref{tab:ops}. Four ablations are reported here and only
three of them rest on observations.

\subsection{Ablation D: equal-suite versus task-weighted}
\label{sec:ablD}

Task weighting raises every score because the larger suites are generally
easier, but it changes no point-estimate position. Equal-suite weighting thus
affects scale, not the cohort ordering, in this version.

\section{Discussion}
\label{sec:discussion}

\subsection{The study needed for run-to-run and engine attribution}
\label{sec:expvariance}

A controlled attribution study remains future work rather than a condition of
the protocol claim, and it is the study that would turn Ablation~A's endpoint
contrast into a route claim. Its preregistered design holds the Qwen3.8-27B checkpoint fixed across five
cells: pinned vLLM BF16, SGLang BF16, vLLM FP8, hosted AkashML BF16, and the
retained self-hosted RunPod endpoint. It reports score, contract-valid
response rate, contract executability and GPU-hours over \ibib-6 Core or a fixed
stratified 64-task sample. Three of the cells require new inference and were not
run here. We therefore make no run-to-run variance, engine, quantization,
batching or hardware attribution claim, and omit the proposed plot rather than
draw interval-free points. The five cells are chosen to resolve exactly the
right-hand column of Table~\ref{tab:observability}, and until the study runs
every endpoint effect in this paper is a joint effect of that column.

\subsection{Saturation, and what the bank still separates (RQ3)}
\label{sec:saturation}

Definition~\ref{def:sat} does identify which suites have stopped separating
frontier systems, and for this instantiation it reports that four of seven are
saturated. That is the RQ3 answer, and Section~\ref{sec:suiteresults} states what
it costs us. The anti-ceiling policy is separately tripped on the same four:
three systems scored exactly 100.00 on S6, four tied at 95.50 on S1, and the
top-six band on S4 spans 94.37 to 97.73. The maintainer completed the required
saturation review on 2026-08-31, confirmed all four triggers, and retained the
preregistered scores and equal-suite weights for v0.13.2, because reweighting a
suite after seeing results would invalidate the frozen comparison; no post-hoc
task, assertion or weight change was made. The signed record is released as
\texttt{paper/data/saturation\_review.csv}.

The reserve that would confirm this against unseen items was consumed for this
cohort (Section~\ref{sec:ethics}), so the four-of-seven result is a
self-diagnosis of the instrument by the instrument: reproducible from the
released suite scores, and not confirmable against items the calibration never
saw. Three of the four suites hold their classification at every band size we
checked and the anti-ceiling policy trips independently on all four, which is why
we do not think it wrong. It is the reason Section~\ref{sec:frozen} makes an
unconsumed reserve a freeze requirement.

The contrast with OfficeQA~Pro sharpens what the finding means. That benchmark
reports frontier systems far lower on enterprise document reasoning over real
collections \citep{opsahlong2026officeqapro} while our S3 leaders sit above 90.
Both instruments cannot be measuring the same construct, and the burden falls on
the one reporting higher scores. The likely explanation is that our packets are
synthetic, thirty pages rather than tens of thousands, and constructed so that
the controlling evidence is always present. We therefore read S3 and S4 as
measuring extraction fidelity under distractors rather than open-collection
document reasoning, and the ``tested setting'' column of Table~\ref{tab:suites}
says so per suite.

One implication cuts against our own corpus contribution. Section~\ref{sec:positioning}
finds \ibib{} to be the only route-bound capability benchmark in our 18-work
audit. This
section reports that the instrument doing that binding has two suites carrying
its composite. Together those say that the contribution which is unique here is
\emph{procedural}, and that the procedure is not specific to our corpus: route
binding, gold-blind adjudication, fail-closed resume and the saturation criterion
could be applied to any bank with a published request contract. C4, the 128-task
instrument, is substantially depreciated by our own analysis of it. The natural
next deliverable is therefore not a bigger private bank but this protocol applied
to somebody else's public corpus. That would answer the private-corpus objection
of Section~\ref{sec:ethics} on its own terms rather than by scoping it, remove
the instrument-fitting row of Table~\ref{tab:threats} because the instrument
would no longer be ours, and test whether the saturation criterion transfers.

\subsection{What the failure taxonomy says about deployment readiness}
\label{sec:failuremeaning}

The failure classes have different operational meanings. Tool-loop exhaustion
indicates a system that plans poorly against a fixed call budget, and it
degrades gracefully in production because the budget is a deployer's choice.
Contract-validity failures are different: they break the machine-readable
interface an enterprise integration depends on. How far downstream tolerance
repairs them depends on the signature. A response that cannot be parsed at all
is unrecoverable by any consumer. A syntactically valid object wrapped in prose,
which is 23 of GLM's 24, is recoverable by a lenient extractor. That is a
failure of \emph{this} contract rather than of every possible one. We report
both under one class because the contract is fixed and published, and flag the
distinction because it changes what an operator should conclude.

\subsection{Path to frozen-v1}
\label{sec:frozen}

v0.13.2 is a candidate calibration, not a certification release. Frozen-v1
requires the controlled remote evaluator, independent two-person gold review,
the ambiguity challenge, signed artifacts, corpus governance and the
prespecified release gate. The analysis above adds five more. The four saturated suites must be
strengthened or their equal contribution reconsidered, and S5 must be reviewed
for construct validity rather than difficulty. The strict-JSON adherence probe
of Section~\ref{sec:probegap} must join the binding gate, blind difficulty
piloting must precede any admission or anti-ceiling decision, and the consumed
challenge reserve must be reconstituted. Three further requirements are
specification or instrumentation rather than new inference. The OCR-capability
predicate must join the gate so the coverage branch becomes reachable,
availability must be reported split by predicate class, and every result record
must carry a zero-inclusive Stage~0 request-attempt count. The present
ten-predicate set now has a dated content-addressed attestation, but it postdates
R12; frozen-v1 requires its successor to be attested before the first locked
launch. Two items the analysis raised are already closed and are
not freeze requirements: the paired intervals resolving the RunPod arm's tier
placement (Section~\ref{sec:ablA}) and the BCa recomputation of the two thinnest
cuts (Section~\ref{sec:sensitivity}). Both were recomputations over retained
per-task scores rather than new inference.

\section{Conclusion}

The serving route is part of the system. A benchmark that scores an advertised
model identifier measures an object nobody deploys. The consequence we defend is
a reporting one: without route binding, capability availability is not a
quantity the result can express, so a number cannot separate a model that failed
from a route that refused. We do not claim to have caught any published
benchmark misattributing one to the other.

The protocol we specify binds the route before scoring, keeps failure in the
score while keeping unsupported capability out of it, and makes recovery
decisions structurally score-blind. Two of its parts are specified more
completely than we ran them. The coverage disposition has no reachable predicate
in this version. And the one route that could test what a \textsc{Bound}
disposition is worth passed all ten predicates, then failed strict-JSON
conformance in production.

Applied to eleven deployed systems, the clearest result is one no
model-identifier benchmark can produce. Two complete calibration runs on routes
serving one set of weights later failed distinct predicates of the finalized
gate, and neither limit was visible from the advertised identifier. The first
surviving signed complete-set attestation postdates those runs, and a passing
receipt predates the fresh Provider~P replacement; this chronology limits the result to descriptive route
evidence rather than preregistered exclusion.
A score-level contrast is also visible: 82.54 against 77.38 under the same
declared upstream revision, precision, and task keys, a $+5.16$-point difference
with 95\% paired interval
$[0.11,10.60]$. We report that as an endpoint effect rather than a route effect,
because the arms also differ in access mode, in harness generation, and in the
serving tool-call parser, and because harness generation is a property of our
evaluator rather than of any endpoint.

The instrument's own results constrain what else it can claim. Four of seven
suites are saturated under a six-system band, three of seven at band five or
seven, and a fifth separates nothing while remaining unsolved. So
we report where systems fail instead of asserting an unsaturated benchmark, and
we report resolution groups instead of ranks. Removing reliability failures from denominators
changes the point ordering, which is why reliability-inclusive scoring is a
substantive choice rather than a wording one. For closed-weight models a
model-only score is not available at all. System-level measurement is the common
ground on which open and closed systems can be compared.

\section{Limitations}
\label{sec:threats}

Table~\ref{tab:threats} states each threat with the direction it biases results,
an estimated magnitude, the mitigation in place, and the residual risk.
Section~\ref{sec:scope} stated the scope before the results; the table states
what each boundary is expected to cost. Four entries carry more of that cost
than the rest.

\paragraph{Vendor conflict of interest.} This one cannot be mitigated by design.
All authors are employed by the party that built the instrument, controlled
scenario design and gold authoring, and holds the reserve-consumption control.
No Iterate system appears in the cohort, but that is not a remedy for
instrument-fitting bias: the bank was built against Iterate systems, and the
eleven third-party systems are its only held-out set. An absent row does not
address it.

\paragraph{Suite saturation.} Four of seven suites are saturated under
Definition~\ref{def:sat}, with band medians from 92.1 to 96.9, so the composite
resolves the frontier less finely than a seven-suite instrument implies. S1's
classification flips at band size five or seven, which is why
Section~\ref{sec:saturation} does not read the count itself as the finding. It
compounds with reserve consumption: the held-out split that could confirm the
classification against unseen items no longer exists in v0.13.2.

\paragraph{Single-run rank instability.} Each configuration was run once, so
run-to-run variance is unidentified and every ordinal claim rests on task-set
resampling alone. The leader pair's interval is 6.4 points wide against a
0.97-point difference. Reporting resolution groups rather than ranks is the
response, and it is a partial one.

\paragraph{The evaluation envelope.} Thirteen images, twenty-five tools and a
completion above 32{,}768 tokens are a design choice of ours rather than a
measured workload population, and they scope every route finding here. That two
of three routes could not execute the contract is a fact about \emph{this}
contract; a different envelope would admit a different set of routes. The first
surviving signed attestation of the complete ten-predicate set postdates the
R11/R12 calibration runs and predates the Provider~P replacement, so only the
latter was demonstrably prospectively bound.

\begin{table}[!t]
\caption{Threats to validity, ordered by our estimate of severity. Direction is
the expected effect on reported scores. Magnitudes are stated as an
order of magnitude with the basis named; where a threat is not quantifiable from
the retained data we say so rather than printing a number.}
\label{tab:threats}
\centering
\footnotesize
\setlength{\tabcolsep}{3.5pt}
\begin{tabular}{@{}p{2.25cm}p{1.75cm}p{2.75cm}p{3.15cm}p{2.45cm}@{}}
\toprule
Threat & Direction & Magnitude & Mitigation in place & Residual risk \\
\midrule
Vendor conflict of interest & upward for framing & not quantifiable: all authors are Iterate employees and control scenario design, gold, and reporting & No Iterate system evaluated; gold frozen before responses; Section~\ref{sec:ethics} & Instrument-fitting bias: the bank was built against Iterate systems and the eleven third-party systems are its only held-out set, so an absent row does not address it; reserve-consumption control is operator-held \\
Suite saturation (S1, S3, S4, S6) & compresses top & 4 of 7 suites by Def.~\ref{def:sat}; band medians 92.1--96.9; S1 flips at band size 5 or 7 & Anti-ceiling policy; reserve split; dated review & Four suites retained for comparability until pre-v1 replacement \\
Reserve consumption $\times$ saturation & unknown sign & removes the held-out split for all 7 suites & Disclosed in Section~\ref{sec:ethics}; classification checked at three band sizes & The 4-of-7 finding cannot be confirmed against unseen items in v0.13.2; unconsumed reserve is a freeze requirement \\
Single-run rank instability & unbiased, high variance & leader-pair task-set interval width 6.4 pts; run variance unidentified & Tiers from all 55 paired task-set intervals & Stochastic ordinal claims unsupported at $n{=}1$ \\
Stage~0 audit asymmetry & unknown relative sign & R17 two-reissue leverage 1.30--2.86 pts; task/suite attribution absent & Aggregate count disclosed; nominal T1--T3 cuts demoted & Ten systems lack zero-inclusive counts; no uniform correction \\
Serving-route dependence & either direction & executability: 2 of 3 routes; score: +5.16 pts, 95\% interval $[0.11,10.60]$ & Route pinning and chronology disclosed; final route preflighted & One run per route \\
Synthetic corpus & upward on S3/S4 & large; S3 leaders $>90$ against a far lower band on OfficeQA Pro & Distractors, decoys, non-adjacent evidence & Does not capture production artifact pathology \\
Private corpus & unfalsifiable by reader & n/a & Retained signed run bundles; synthetic worked items (App.~\ref{app:worked}) & Item-level inspection unavailable \\
Tool sandbox abstraction & upward on S6 & large; S6 top-six median 96.9 & Bounded per-task catalog, stateful execution & Not a customer's production tools \\
Equal-suite weighting & shifts small suites up & see Sec.~\ref{sec:ablD} & Task-weighted diagnostics retained & Design choice, not derived \\
Cost scope & downward & excludes infra & Ledger-observable usage, run-accounting snapshot & Not total cost of ownership \\
No human baseline & no anchor & n/a & None & Absolute scores lack business meaning \\
Evaluation envelope & scopes all route findings & 13 images / 25 tools / ${>}32{,}768$ tokens, not derived from a workload population & Contract published in full (Table~\ref{tab:contract}) & Route findings hold under this contract only \\
Organizational scope & unknown & not quantifiable: no workload trace or multi-firm survey; 12 families were selected by one firm & Twelve scenario families & Reflects one firm's scenario design \\
\bottomrule
\end{tabular}
\end{table}

\section{Ethics, conflicts of interest, and responsible release}
\label{sec:ethics}

Four of the disclosures below are load-bearing rather than procedural: the
conflict of interest, the consumed reserve, the gold-integrity attestation, and
the unfalsifiability of a private corpus.

\paragraph{Affiliation and commercial interest.} All authors are employed by
Iterate.ai, which sells enterprise AI systems. This paper evaluates and ranks
eleven third-party systems, several of which are sold by competitors.

\paragraph{No Iterate system was evaluated.} None of the eleven systems in
Table~\ref{tab:systems} is an Iterate product or contains an Iterate component.
The benchmark was built as an internal release gate and no internal result
appears in this paper.

\paragraph{Funding and inference cost.} Iterate.ai funded the study and paid
the full measured inference cost of \$245.99 in Table~\ref{tab:ops}. No model
vendor, serving provider, or aggregator supplied research credits, discounted
access, free inference, or early access for this study.

\paragraph{Vendor notification and right of reply.} No evaluated vendor was
consulted about the protocol, notified before the runs or manuscript, or given
a pre-publication right of reply. Consequently, no vendor response was
incorporated.

\paragraph{Reserve consumption.} We consumed the reserve for this cohort.
The retained configurations cover all seven suites, set
\texttt{challenge\_reserve\_acknowledged} to true, and record the literal
acknowledgment \texttt{consume-private-reserve}. The first retained
complete-candidate run began consuming the all-suite bank at
2026-07-23 05:50:04 UTC. The accountable operator role, attested by the authors,
was the \ibib{} Benchmark Maintainer. The control was held by an Iterate employee.

\paragraph{Gold integrity.} A gold answer or scoring rule is never changed after
observing a vendor response. The \ibib{} Benchmark Maintainer is accountable for
gold integrity. The attestation mechanism is the versioned corpus manifest and
its SHA-256 lock: \texttt{make lock-verify} must pass before evaluation, and a
material corpus or gold change requires a version bump, a resealed manifest,
and a recorded change review. Published run plans bind the corpus-manifest
digest, so an unrecorded change breaks comparison.

\paragraph{Private evaluation and its risks.} A private corpus is
unfalsifiable by the reader, and the risks of evaluation by private data
curators are documented \citep{bansal2025peeking}. We do not regard our governance
machinery as a sufficient answer to that argument. The split of
Section~\ref{sec:naming} changes the scope of the objection rather than its
force: the protocol is published and can be contested, re-implemented, or
applied to a corpus we never see, whereas the reference instantiation and the
eleven-system cohort reported here inherit the objection in full. A reader who
rejects our numbers on this ground can still adopt, or refute, the procedure
that produced them. We regard three disclosures
as the minimum: worked synthetic conformance items covering every suite
(Appendix~\ref{app:worked}), the realized assertion and difficulty distributions
(Appendix~\ref{app:assertions}), and the run-disposition ledger
(Table~\ref{tab:rundisposition}).
Precedent for credible private evaluation exists in SEAL and the ARC Prize
\citep{scale2024seal,chollet2024arcprize}, and both disclose more than the
present artifact does.

\paragraph{Data availability.} The methodology, schemas, runner interfaces,
synthetic worked items, run-plan manifest field list, verification tooling, and
the dated audit ledgers and derived tables behind every table in this paper
will be released at \url{https://github.com/IterateAI/IBIB}. The release is a
deny-by-default export produced by \texttt{make verify}; each exported file is
content-addressed in \texttt{MANIFEST.sha256}. The public release date is
defined as the UTC creation timestamp of the reviewed release tag. Code and
verification tooling are licensed under Apache-2.0;
the manuscript, tables, and figures are licensed under CC BY 4.0. The private
corpus, task bank, gold assertions, and non-public run evidence are expressly
excluded from both grants and remain behind a controlled evaluator.

\paragraph{Permission to publish named comparative results.} This paper names
eleven commercial systems, routes several of them through an aggregator, and
publishes a table in which some of them score below 50. Several provider
agreements restrict benchmarking of their services. On 2026-08-31 and
2026-09-07 the authors reviewed the public terms for OpenAI, Anthropic,
OpenRouter, every pinned upstream provider, and both named direct infrastructure
routes. The dated provider-by-provider record and URLs are released
as \texttt{paper/data/terms\_audit.csv}. We identified no explicit prohibition
on publishing aggregate comparative research for the named routes other than
Provider~P, whose public terms expressly prohibit benchmarking the platform or
services. We therefore withhold that upstream identity, publish no raw output,
make no claim about its provider-level performance, and require written
permission before naming it. Model and service identifiers elsewhere are
factual experimental labels and do not imply vendor participation, endorsement,
or a commercial relationship. Private order forms may supersede public terms;
this record is a responsible-release control, not legal advice.

\paragraph{AI-assisted writing.} OpenAI Codex and ChatGPT assisted with code
review, evidence extraction, manuscript editing, consistency checks, and figure
generation. The human authors selected the claims, reviewed the source evidence,
verified the reported numbers and citations, and accept responsibility for the
paper. No AI system is listed as an author.

\paragraph{Human subjects.} No human subjects were involved in the present
study.

\clearpage
\bibliographystyle{plainnat}
\bibliography{references}

\clearpage
\appendix

\clearpage

\section{Protocol algorithms and their dispatch tables}
\label{app:algorithms}

\paragraph{The classification step.}
The result schema admits exactly five statuses (\texttt{ok}, \texttt{error},
\texttt{timeout}, \texttt{invalid\_output}, and \texttt{unsupported}) and three
coverage states (\texttt{supported\_attempted}, \texttt{unsupported}, and
\texttt{not\_run}). Algorithm~\ref{alg:adjudicate} is exhaustive over the causal
classes derived from those values and the sanitized provider envelope.

Table~\ref{tab:classify} is $\textsc{Classify}$. Naming the classes and calling
them a grouping of the ledger's exhaustive causal subclasses says what they
mean but not how an independent group would assign one, and
$\textsc{Classify}$ is the crux of the gold-blindness claim: it is the single
step between a raw failure and a scoring consequence. The table is evaluated in
order and the first matching row wins, so the tie-break is the row order and
nothing else. Its inputs are exactly the projection $\hat{x}$: status,
coverage state, and a sanitized provider-envelope summary. The summary is
derived from the raw response before any grader runs. The implementation
constructs a new dictionary with exactly the allowlisted fields before
classification; it does not pass a view of the scored record, so gold-derived
fields are unreachable by type and construction. The terminal row makes the
table total: any $\hat{x}$ that matches nothing above it is
\emph{unclassified}, which routes to \textsc{Hold}, so no unmodelled failure is
ever silently scored.

\subsection{Scoring}

Let assertion credit $a_{ij} \in [0,1]$ be the deterministic credit earned by
task $i$ on assertion $j$ with predeclared weight $w_{ij}$. The task score is
\begin{equation}
T_i = 100 \cdot \frac{\sum_j w_{ij} a_{ij}}{\sum_j w_{ij}}.
\end{equation}
For suite $k$, $S_k$ is the arithmetic mean of its locked task scores. \ibib-7
Full is $\frac{1}{7}\sum_{k=1}^{7} S_k$ and \ibib-6 Core is the equal-weight
mean over S1, S2, S3, S5, S6, and S7. Equal-suite weighting prevents the
22-task suites from dominating the smaller S1 and S7. Task-weighted means are
retained as diagnostics and are reported against the equal-suite ranking in
Section~\ref{sec:ablD}.

The primary measure is the first-pass, reliability-inclusive, equal-suite
\ibib-7 Full score. The v0.13.2 implementation has \emph{no causal-class score
constants and no run-plan scoring block}. Every attempted result is passed once to the same deterministic
assertion grader. A transport, timeout, runtime, empty-final, or other failure
with no parseable response earns zero because all declared assertions fail. A
parseable but contract-invalid response can earn non-zero assertion credit for
the work products it actually supplies; its causal class does not add or
subtract points. Tool-budget and tool-loop outcomes are treated the same way:
the retained final work product, if any, is graded, and an absent final work
product earns zero. Thus the only fixed value is the mechanically implied zero
for an absent response, not a class-dependent penalty. The signed plan binds the
scorer version, task and gold hashes, and assertion inventory. An unsupported
task is not attempted and is never converted to a zero. A complete-profile
composite requires every task in every required suite to have been attempted.

The assertion weights $w_{ij}$ are likewise predeclared, and predeclaration is
a timing property rather than a procedure, so we state the procedure. Weights
are assigned when a task is authored, and they sum to 100 points. They come
from the ``required work product'' and ``evaluated result'' fields of that
task's row in Table~\ref{tab:suites} (Section~\ref{sec:workrep}), and they are
fixed before any system response for that task exists. For S1, source evidence receives 10 points and the
remaining 90 are divided equally among the declared scalar answer fields. For
the core S2--S5 templates, source evidence receives 25 and the remaining 75 are
divided equally; the four frontier-extension families in each of S3--S5 use a
20/80 split instead. No scalar field, including a controlling numeric value,
receives a hidden bonus: fields within a template have equal weight. S6 fixes
the required work-product vector at 40 points for tool trace, 20 for final
state, and 10 each for outcome, action count, recovery behaviour, and evidence.
S7 assigns 42 points equally across its twelve scalar fields (3.5 each), 22 to
the exception ledger, 16 to the period bridge, 10 to source evidence, and 10 to
SQL-trace policy. These rules generate every allocation in the locked assertion
inventory, and they are checkable against the realized assertion counts of
Table~\ref{tab:assertions}: they predict exactly the six assertions per task
that S6 carries and the sixteen that S7 carries. They are \emph{not} the
allocations shown in Appendix~\ref{app:worked}. Those fixtures are synthetic
cases authored after the runs to exercise the scoring predicates. They carry
three to five assertions where the sealed tasks carry five to sixteen. Their
point splits are chosen to make a specific predicate legible rather than to
instantiate the authoring rule. The rule above is what a third party
reimplementing the protocol should apply; the appendix shows how the scorer
behaves once weights, whatever their provenance, are fixed.

Token efficiency is a diagnostic and is not folded into the headline. For a
complete run, let $C$ be the sum of provider-reported completion tokens across
all task turns and let $Q = \sum_i T_i / 100$ be the task-weighted number of
fully-correct-task equivalents. We report $10^6 Q/C$. Failed attempts contribute
zero to the numerator while their tokens remain in the denominator.

\section{Instrument and cohort detail}
\label{app:instrument}

\subsection{Cohort}

Table~\ref{tab:systems} lists the eleven evaluated systems. The cohort is
purposive rather than exhaustive. Three GPT-5.6 endpoints measure one provider
family at its published \texttt{xhigh} setting across three operating tiers
\citep{openai2026gpt56}. Claude Fable~5 and Claude Sonnet~5 give two current Anthropic
operating points through the direct Messages API
\citep{anthropic2026fable5,anthropic2026sonnet5}. Qwen3.5-122B-A10B provides a
commercially served sparse comparison \citep{qwen2026qwen35}; Qwen3.6-27B a
smaller dense one \citep{qwen2026qwen36}; Qwen3.8-27B the current dense flagship
\citep{qwen2026qwen38}. Kimi K3 adds a distinct multimodal tool-capable
architecture \citep{moonshot2026kimik3}, and Muse Glimmer 30B an always-on
local-agent-class open-weight comparison \citep{meta2026museglimmer}. The GLM
entry composes GLM-5.2 for text and tool suites with GLM-5V-Turbo for visual
suites under one preregistered identity
\citep{zai2026glm52,zai2026glm5vturbo}.

This is the cohort frozen before execution. We use each evaluated system's
official published display name; a later release is not substituted for a
scored row without a fresh bound 128-task run.

Every open-weight system in the primary cohort uses exactly one serving route,
per Property~\ref{prop:nonsub}. OpenRouter routes must be zero-data-retention
eligible and must deny data collection, and the returned model and upstream
provider values are retained for every response.

\subsection{What is observable and what is not}

Table~\ref{tab:systems} records $m$, $e$, $p$ and $r$ because those are
recoverable from a hosted route. Several determinants of behaviour are not.
Table~\ref{tab:observability} separates them, and the right-hand column is the
argument for why a reference-deployment axis must eventually exist alongside the
served-route axis (Section~\ref{sec:serving}). Every entry on the right is known
to move benchmark accuracy for identical weights
\citep{pape2026silenthyperparameter,kurtic2025bf16}, which is why we name them
rather than treat them as infrastructure.

\subsection{Execution, cost, and latency accounting}

Tasks are issued serially at concurrency one. Latency is evaluator-observed per
task and includes provider queueing, evaluator-owned PDF and SQL work, and
multi-turn tool use; it is not isolated model-only inference latency. Cost is a
measured run diagnostic rather than a reconciled invoice. Provider-hosted
systems use ledger-observable returned usage; OpenRouter cost is the sum of
provider-reported per-response cost; OpenAI cost applies a timestamped official
price snapshot to provider-reported usage, including cache writes at the
published $1.25\times$ input rate, cached reads, output tokens, and the
published long-context multiplier; the snapshot revision, its digest, and the
per-family rate table are in Appendix~\ref{app:cost}. Missing or internally inconsistent usage
places the cost analysis on hold, and an unsuccessful attempt returning no usage
record causes the observed sum to be labelled a lower bound and the system to be
excluded from cost-Pareto membership.

The sampling parameter set was frozen before the first cohort request and is
printed in Appendix~\ref{app:manifest}. The retained manifests do not record the
exact freeze event; they prove that it occurred no later than
2026-07-23 05:50:04 UTC. That instant is also when the first retained
complete-candidate run began (Section~\ref{sec:ethics}), so the evidence
establishes only that the parameters were frozen by the time the first task was
sent. It is a bound derived from the first run, not an independent
preregistration receipt, and we do not present it as one; a dated freeze
attestation separate from the first launch is a freeze requirement
(Section~\ref{sec:frozen}). Every row requested 65{,}536 output tokens, and no
route exposed a reproducible sampling seed. Provider-default controls are
reported as unexposed rather than reverse-engineered.


\subsection{Contamination control and the challenge reserve}
\label{sec:contamination}

The corpus, gold assertions, and reserve split remain private. To make privacy
compatible with auditability, the candidate lock inventories every task, gold
record, artifact, evaluator module, schema, and taxonomy file by SHA-256 and
size. The evaluator may be resealed only when the locked corpus content is
byte-for-byte unchanged, and the new manifest must record both its predecessor
hash and the reason for resealing.

Approximately 30\% of each suite forms a challenge reserve whose purpose is
saturation detection, not iterative prompt tuning. A complete candidate
calibration may consume the reserve only after an operator gives explicit
acknowledgment. Because that operator is employed by the party publishing the
benchmark, we report reserve consumption status for this cohort in
Section~\ref{sec:ethics} rather than leaving the control undisclosed.

\subsection{Suite construction}
\label{sec:suiteconstr}

S1 uses synthetic single-tab workbooks with explicit business rules, distractor
controls, and auditable source ranges. S2 adds multi-tab joins, temporal
controls, pivot-table behavior, stale caches, and duplicate reconciliation. S3
packets contain thirty pages of selectable-text tables and require evidence from
non-adjacent pages. S4 is image-only, so extracted PDF text is never supplied.
S5 combines charts, organizational structures, architecture paths,
critical-chain schedules, and risk matrices. For native vision adapters, S4 and
S5 render to deterministic JPEG pages at 180 DPI, and neutrality tests verify
that providers receive byte-identical content in the same pixel order.

S6 draws on an evaluator-side registry of more than one hundred enterprise
tools, and exposes only the bounded relevant subset per task. This avoids two
artificial extremes: a global tool set small enough to make selection trivial,
and a hundred-tool prompt dominated by irrelevant options. The system selects a
tool and its arguments; the evaluator executes the call inside a stateful
sandbox. Scoring covers selection, order, arguments, unnecessary calls,
recovery, the final answer, side-effect safety, and the audit trail. One call is
admitted per turn, no task exceeds twelve calls, and every task requires a
final-answer turn. Provider parallelism cannot bypass this evaluator-owned
budget.

S7 uses a physical SQLite database with exactly ten million fact rows and a
populated 1{,}000-column relation. The database is never placed in model
context. The system receives a compact relation catalog and access to a
governed, read-only SQL tool. A field dictionary maps business terms to
certified physical columns hidden among decoys and superseded names; raw schema
metadata and unrestricted PRAGMA access are blocked. Schema-discovery credit is
awarded only when a successful query returns the certified relation, business
term, and physical column together. SQL is bounded by text length, projection
width, result rows, cell size, response size, and execution time.

\subsection{Task model}
\label{sec:taskmodel}

Every task is defined by an immutable task identifier, benchmark version, suite,
scenario family, and difficulty label; capability tags and permitted failure
labels; one or more hashed inputs; a provider-visible prompt with business
rules; a typed output contract; citation requirements; deterministic gold
assertions with numeric tolerances; and a privacy classification. Task files and
gold files are physically separate, so gold data cannot enter a provider-visible
request.

The v0.13.2 bank spans sales, accounts payable, inventory, customer success,
claims, procurement, workforce, logistics, compliance, market intelligence,
operational troubleshooting, and forensic accounting.

\subsection{Difficulty, admission, and the anti-ceiling policy}

The bank targets a mixed difficulty profile of approximately 20\% Anchor, 40\%
Hard, 30\% Expert, and 10\% Frontier-breaker. Labels remain provisional until
blind pilot evidence supports them, and the realized distribution is reported in
Appendix~\ref{app:assertions} rather than assumed from the target. That
distribution is worse than mis-targeted, and we state the stronger finding
rather than only the weaker one. The realized labels are effectively constant
within a suite. S2, S3, S4 and S5 carry byte-identical 18/4
Expert/Frontier splits, so they carry no within-suite information at all, and
those four suites span the entire measured range of the bank, from
discriminative (S2, $\rho = 12.5$) to saturated (S3 and S4, $\rho = 1.8$ and
1.7). With 93.8\% of tasks labelled Expert or Frontier while four of seven
suites are saturated, the labels are not merely mis-distributed relative to the
target; they are non-predictive of measured difficulty. Nothing in this paper's
results rests on them, but the admission gate and the anti-ceiling policy below
both reference difficulty, so we treat the taxonomy as an unvalidated construct
in v0.13.2 and make blind difficulty piloting a freeze requirement in
Section~\ref{sec:frozen} rather than defending the labels here.

Difficulty must come from the work. Every task must be answerable from the
supplied evidence and permitted tools. Ambiguous wording, illegible inputs,
undisclosed conventions, and adversarial grading are not valid sources of
difficulty. Candidate admission requires task and gold consistency, reference
conformance, artifact integrity, schema validation, and deterministic regrading.
Frozen-v1 admission adds independent two-person gold review and an ambiguity
challenge; local automated tests cannot substitute for those human attestations.

The anti-ceiling policy does not cap a score. A correct answer receives full
credit even in the nineties. Instead, a repeated later-suite mean above 95
triggers a saturation review, and a repeated mean above 97.5 prevents that suite
from being frozen. Any revision must introduce new or stronger authentic
enterprise constructs in a new benchmark version. A gold answer or scoring rule
is never changed after observing a vendor response.

The trigger above is a governance rule about a single mean. For reporting we
need something that survives one aberrant system in a six-system band, because
neither the arithmetic mean nor the range over six points does. We therefore fix
one operational definition and use it everywhere in this paper.

Three properties of this definition are deliberate. It is checked in order, so a
suite at ceiling is never called discriminative because one system collapsed on
it. Every quantity is a selection from the sorted band rather than an
interpolation, so no quantile convention is involved: on this cohort the
interquartile range of S6 is 9.1 or 14.3 depending on the convention chosen, and
that choice alone would decide whether the bank has two separating suites or
three. And the trimmed range discards one system at each end, which is the
smallest robustness that answers the objection the untrimmed range invites.

The band size is the one free parameter here, and it is fixed at six rather
than derived, so we report what the classification does at neighbouring sizes.
Holding the $\geq 90$ count at the same two-thirds proportion of the band
($\lceil 2n/3 \rceil$), the classification of five of the seven suites is
invariant across band sizes five, six and seven: S3, S4 and S6 are saturated at
every size, S2 is discriminative at every size, and S5 is compressed below
ceiling at every size. Two move. S1 is saturated at six ($n_{90}=4$ of 6) and
compressed below ceiling at five and seven ($3$ of 5, $4$ of 7); it is never
discriminative at any size. S7 is discriminative at six and seven and
compressed at five, where the trimmed range over a five-system band collapses
to 2.1. So the ``four of seven saturated'' headline is a six-band statement and
would read ``three of seven'' at five or seven, with S1 the only suite in
question; the claim that the bank has stopped separating on S3, S4 and S6, and
that S5 separates nothing, holds at every band size we checked. This
sensitivity is also the reason we do not read the count itself as the finding.

The classification is against our own interest in both directions. The
untrimmed range would let us report S6 as discriminative on the strength of a
single system, and a mean-based ceiling test would let us report S6 and S1 as
unsaturated. We report four saturated suites.
Section~\ref{sec:saturation} reports the status of the policy and of
Definition~\ref{def:sat} against the present cohort, including the suites where
both are tripped.

\begin{table}[t]
\caption{Observability of the system tuple on a hosted route.}
\label{tab:observability}
\centering
\small
\begin{tabular}{@{}p{6.2cm}p{6.2cm}@{}}
\toprule
Observable and recorded & Structurally unobservable on hosted routes \\
\midrule
Model identity and pinned revision; upstream provider; served precision where
disclosed; token usage; finish reason; per-response cost; observed completion,
image, and tool envelopes; data-retention posture
&
Inference engine and version (vLLM, SGLang, TensorRT-LLM); guided-decoding
backend (xgrammar, outlines, lm-format-enforcer); tool-call parser and chat
template version; KV-cache dtype, prefix caching, chunked prefill, speculative
decoding; hardware generation \\
\bottomrule
\end{tabular}
\end{table}

\subsection{Assertion schema and realized distributions}
\label{app:assertions}

Each gold record contains a task identifier, benchmark version, maximum points,
a digest of the gold answer, and a list of typed assertions. Every assertion
has an identifier, type and positive point value. The realized types are:
\texttt{exact} (279), \texttt{numeric} (508),
\texttt{citation\_contains} (10), \texttt{citation\_match} (118),
\texttt{table} (24), \texttt{tool\_trace} (18),
\texttt{tool\_recovery} (18), and \texttt{sql\_trace\_policy} (12).
Points are weights: an assertion's awarded points are divided by the task's
\texttt{max\_points}, then multiplied by 100. Exact assertions compare
case-folded whitespace-normalized values. Numeric and numeric table cells use
\texttt{math.isclose} with the assertion's predeclared absolute and relative
tolerances (default zero). Partial credit is available only when explicitly
enabled: set/SQL/table assertions use the matched-item or matched-cell ratio,
and tool traces use the fraction of required ordered calls matched.

Every task requires citations and has exactly one citation assertion.
\texttt{citation\_contains} requires an artifact and sheet match plus coverage
of each required A1 range. \texttt{citation\_match} requires every declared
locator to be contained, key for key, in at least one returned citation;
unlisted citation keys do not affect the match.

\begin{table}[!ht]
\caption{Realized assertion counts. Medians are assertions per task. The total
is the sealed-bank aggregate; no expected value or task content is disclosed.}
\label{tab:assertions}
\centering\small
\begin{tabular}{@{}lrrrrr@{}}
\toprule
Suite & Tasks & Assertions & Min & Median & Max \\
\midrule
S1 & 10 & 51  & 5  & 5  & 6 \\
S2 & 22 & 162 & 7  & 7  & 9 \\
S3 & 22 & 154 & 7  & 7  & 7 \\
S4 & 22 & 158 & 7  & 7  & 8 \\
S5 & 22 & 162 & 7  & 7  & 9 \\
S6 & 18 & 108 & 6  & 6  & 6 \\
S7 & 12 & 192 & 16 & 16 & 16 \\
\midrule
Total & 128 & 987 & -- & -- & -- \\
\bottomrule
\end{tabular}
\end{table}

\begin{table}[!ht]
\caption{Realized task difficulty labels. These do \emph{not} match the planned
20/40/30/10 Anchor/Hard/Expert/Frontier-breaker mix: the sealed records use
Intermediate and Advanced in S1, no Anchor or Hard label appears, and S6/S7 are
single-band. We report the mismatch rather than map labels post hoc. The
informational finding is stronger than the distributional one: outside S1 every
suite is single- or two-valued, S2--S5 carry byte-identical 18/4 splits despite
spanning the full measured range of the bank, and the labels are therefore
suite-constant and non-predictive of measured difficulty. No result in this
paper is derived from them; Section~\ref{sec:frozen} makes blind difficulty
piloting a freeze requirement.}
\label{tab:difficulty}
\centering\small
\begin{tabular}{@{}lrrrrr@{}}
\toprule
Suite & Intermediate & Advanced & Expert & Frontier & Total \\
\midrule
S1 & 2 & 6 & 2 & 0 & 10 \\
S2 & 0 & 0 & 18 & 4 & 22 \\
S3 & 0 & 0 & 18 & 4 & 22 \\
S4 & 0 & 0 & 18 & 4 & 22 \\
S5 & 0 & 0 & 18 & 4 & 22 \\
S6 & 0 & 0 & 18 & 0 & 18 \\
S7 & 0 & 0 & 0 & 12 & 12 \\
\midrule
Total & 2 & 6 & 92 & 28 & 128 \\
Percent & 1.56 & 4.69 & 71.88 & 21.88 & 100.00 \\
\bottomrule
\end{tabular}
\end{table}


\section{Operating characteristics, run disposition, ablations and intervals}
\label{app:ops}

\subsection{Run disposition}
\label{sec:rundisp}

The publication lineage contains seventeen full-cohort launches or partial
cohort launches, of which eleven are published, four are superseded by fresh
whole-route runs, and two redundant launches were stopped before completion.
No response was spliced across those full-run replacements. The complete ledger,
including full run identifiers, UTC timestamps, gold-blind reasons, observed
scores, the four replacement deltas, and the available Stage~0 audit status, is released as
\texttt{paper/data/run\_dispositions.csv}. Synthetic binding probes, diagnostic
subsuite studies, and task-level fixed-cap recovery attempts are not independent
cohort launches; they remain enumerated in the signed preflight receipts and
recovery registry.


\begin{table}[t]
\caption{Operating characteristics. \emph{Coverage} here means capability-lane
support: every system was admitted on all seven lanes by
Algorithm~\ref{alg:bind}, so no lane is withheld and no coverage column is
shown. Lane admission is a binding-time property and is distinct from the
production contract failures recorded for GLM in Section~\ref{sec:glm}. This is a different quantity
from the \emph{Valid} column of Table~\ref{tab:main}, which counts tasks that
returned a scoreable response. $Q/C$ is fully-correct-task equivalents per
million completion tokens; tokens spent on failed attempts remain in the
denominator.}
\label{tab:ops}
\centering
\small
\setlength{\tabcolsep}{4pt}
\begin{threeparttable}
\begin{tabular}{@{}lrrrrrr@{}}
\toprule
System & Cost (USD) & Med.\ lat.\ (s) & p95 lat.\ (s) & Tok./task & $10^6Q/C$ & Reas.\ share (\%) \\
\midrule
GPT-5.6 Sol       & 36.93 & 36.3  & 330.0  & 5{,}239  & 171.81 & 84.7 \\
Claude Fable 5    & 89.02 & 35.3  & 262.5  & 5{,}567  & 159.53 & 77.9 \\
GPT-5.6 Terra     & 18.27 & 18.3  & 240.3  & 5{,}123  & 170.16 & 86.8 \\
Qwen3.8-27B       &  9.26 & 199.0 & 990.1  & 17{,}193 &  48.94 & 57.6 \\
GPT-5.6 Luna      &  8.39 & 30.7  & 197.2  & 6{,}968  & 117.64 & 95.3 \\
Claude Sonnet 5   & 27.38 & 49.7  & 306.7  & 11{,}372 &  68.74 & 90.1 \\
Muse Glimmer 30B  &  2.43 & 115.9 & 1119.2 & 10{,}159 &  75.43 & n/r \\
Kimi K3           & 26.19 & 47.0  & 222.8  & 9{,}173  &  82.08 & 67.2 \\
Qwen3.6-27B       &  7.07 & 267.8 & 1046.4 & 14{,}938 &  48.51 & 53.5 \\
Qwen3.5-122B-A10B &  5.70 & 85.2  & 326.7  & 10{,}947 &  50.71 & 96.2 \\
GLM-5.2 + GLM-5V-Turbo & 15.35 & 106.8 & 607.0  & 19{,}334 &  22.46 & 96.9 \\
\midrule
\textbf{Cohort}   & \textbf{245.99} & & & & & \\
\bottomrule
\end{tabular}
\begin{tablenotes}[flushleft]\footnotesize
\item n/r: reasoning tokens not separately reported by the provider. The
unrounded aggregate ledger sums to \$245.9914; the displayed cohort total is
therefore \$245.99.
\end{tablenotes}
\end{threeparttable}
\end{table}

\begin{table}[t]
\caption{Complete disposition denominator for the v0.13.2 publication lineage.
The row keys resolve to full run identifiers in
\texttt{paper/data/run\_dispositions.csv}. Deltas are replacement minus
superseded score and are descriptive because each route was run once.}
\label{tab:rundisposition}
\centering\small
\setlength{\tabcolsep}{2.5pt}
\begin{threeparttable}
\begin{tabular}{@{}p{1.5cm}p{4.1cm}p{2.0cm}p{3.8cm}r@{}}
\toprule
Rows & System & Disposition & Gold-blind basis & $\Delta$ \\
\midrule
R01--R03,R05 & GPT-5.6 Sol, GPT-5.6 Terra, Qwen3.5-122B-A10B, GPT-5.6 Luna & published & complete signed runs & -- \\
R04$\to$R14 & Kimi K3 & superseded & unstable response path; fresh route & +47.93 \\
R06--R07 & GPT-5.6 Sol, Qwen3.5-122B-A10B & excluded partial & redundant launches; no score consulted & -- \\
R08--R10 & Claude Fable 5, Claude Sonnet 5, GLM-5.2 + GLM-5V-Turbo & published\tnote{a} & complete signed runs & -- \\
R11$\to$R13 & Qwen3.6-27B & superseded & complete run; later image-limit audit & +42.14 \\
R12$\to$R13 & Qwen3.6-27B & superseded & complete run; later long-output audit & +6.01 \\
R15 & Muse Glimmer 30B & published & complete signed run & -- \\
R16$\to$R17 & Qwen3.8-27B & superseded & preregistered economics correction & +5.16 \\
\bottomrule
\end{tabular}
\begin{tablenotes}[flushleft]\footnotesize
\item[a] The GLM run is published as a served-route diagnostic whose S6 and S7
scores are qualified as measured contract failures of the served route. They
are \emph{not} a \textsc{PartialCoverage} disposition: that branch of
Algorithm~\ref{alg:bind} never fired for this route, nor, per
Section~\ref{sec:binding}, for any route in this cohort. No lane was
withheld, and both suites are scored and included in the composite
(Section~\ref{sec:glm}). R11 and R12 are separate complete 128-task runs of the
same weights on two routes; both were superseded, so each has its own descriptive
delta to R13. The finalized gate postdates both, as attested in
\texttt{paper/data/binding\_predicate\_freeze.csv}; no response was spliced.
\end{tablenotes}
\end{threeparttable}
\end{table}

\begin{figure}[t]
\centering
\includegraphics[width=\textwidth]{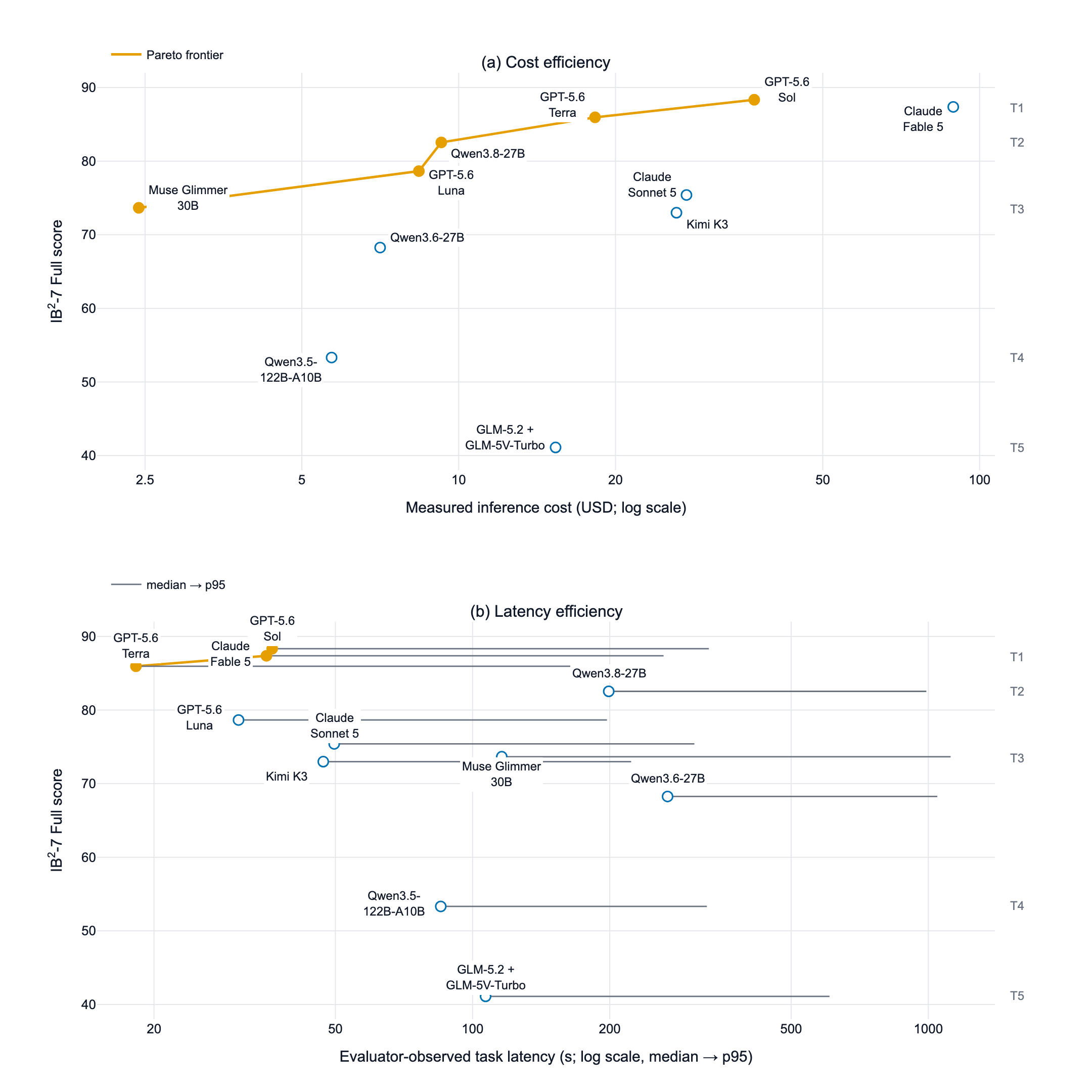}
\caption{Score against measured inference cost (top) and evaluator-observed task
latency (bottom). Filled points and orange lines mark the empirical Pareto
frontiers, computed on cost and on \emph{median} latency respectively. Grey
horizontal bands are the nominal tier diagnostics of Table~\ref{tab:main}, labelled at
the right: the connected frontier is a cost or latency ordering and is not the
score ordering, which Section~\ref{sec:tiers} declines to report at $n{=}1$, and
the bands are there so the two are not confused. In the lower panel each system
carries a whisker from its median to its p95, because a median-only latency
axis flatters the two systems whose p95 is near 1{,}000 seconds. Both axes are
log-scaled. All eleven cost ledgers are complete and none is a lower bound.}
\label{fig:efficiency}
\end{figure}

\subsection{Ablation tables}

\begin{table}[t]
\caption{Route-lineage ablation for identical Qwen3.6-27B weights. R11 and R12
were complete locked calibration runs; their capability limits were classified
only after the finalized gate was attested. Provider~P passed that gate before
its fresh replacement run. Scores are descriptive single-run contrasts, not
inferential route effects.}
\label{tab:ablroute}
\centering\small
\begin{tabular}{@{}llll@{}}
\toprule
Route & Locked run timing & Finalized-gate outcome & Locked score \\
\midrule
DeepInfra FP8 & R11; before gate freeze & predicate 4: 13 images rejected; limit 4 & 26.12 \\
CoreWeave FP8 & R12; before gate freeze & predicate 7: length at 32{,}768 & 62.25 \\
Provider P FP8 & R13; after gate freeze & pass; 52{,}454 tokens, stop & 68.27 \\
\bottomrule
\end{tabular}
\end{table}

\begin{table}[t]
\caption{Component-by-component difference between the two Qwen3.8-27B arms of
Ablation~A, in the tuple of Definition~\ref{def:sut}. Property~\ref{prop:nonsub}
makes any differing component sufficient to distinguish two systems, so the
table is what licenses, or withholds, an attribution of the 5.16 points to
$r$ alone. The envelope, tool and output-contract rows are resolved from the
two signed run plans.}
\label{tab:armdiff}
\centering\small
\setlength{\tabcolsep}{4pt}
\begin{threeparttable}
\begin{tabular}{@{}llll@{}}
\toprule
Component & R16 (77.38) & R17 (82.54) & Same? \\
\midrule
$m$ model identity & Qwen3.8-27B & Qwen3.8-27B & yes \\
$p$ precision & BF16 & BF16 & yes \\
$e$ reasoning effort & xhigh & xhigh & yes \\
$r$ serving route & direct RunPod & OpenRouter $\to$ AkashML & \textbf{no} \\
$H$ harness generation & fair-v5 & fair-v6 & \textbf{no}$^{\dagger}$ \\
$\Omega$ envelope & $(262{,}144,65{,}536,13,12)$ & same & yes \\
$\mathcal{T}$ tool implementation & vLLM \texttt{qwen3\_coder} parser & AkashML Qwen parser & \textbf{no}$^{\ddagger}$ \\
$\mathcal{C}$ output contract & strict JSON / evaluator validation & same & yes$^{\S}$ \\
\midrule
Task keys and weights & \multicolumn{2}{l}{identical} & yes \\
Transport correction & none & none & yes \\
\bottomrule
\end{tabular}
\begin{tablenotes}[flushleft]\footnotesize
\item $^{\dagger}$Equivalent on the enumerated surface of
Section~\ref{sec:results}'s bridge, not byte-identical.
\item $^{\ddagger}$The evaluator tool catalog, SQL tool and schema-discovery
contract hashes are identical, but the serving parser differs, so
$\mathcal{T}$ is not byte-identical.
\item $^{\S}$The canonical response-contract, structured-output and
tool-validation fields are byte-identical.
\end{tablenotes}
\end{threeparttable}
\end{table}

\begin{table}[t]
\caption{Reliability-inclusive primary score versus conditional accuracy over
contract-valid responses. Conditional accuracy equal-averages the seven
within-suite valid-response means. GLM is withheld because S7 has no valid
response; averaging six suites would change the profile. Positions are point-
estimate diagnostics, not new evidence tiers.}
\label{tab:ablaccuracy}
\centering\footnotesize
\begin{tabular}{@{}lrrrr@{}}
\toprule
System & First pass & Accuracy only & $\Delta$ & Accuracy-only position \\
\midrule
GPT-5.6 Sol & 88.34 & 89.03 & +0.69 & 2 \\
Claude Fable 5 & 87.37 & 87.37 & +0.00 & 3 \\
GPT-5.6 Terra & 85.95 & 86.64 & +0.69 & 5 \\
Qwen3.8-27B & 82.54 & 83.67 & +1.13 & 6 \\
GPT-5.6 Luna & 78.65 & 90.10 & +11.45 & 1 \\
Claude Sonnet 5 & 75.40 & 76.03 & +0.63 & 9 \\
Muse Glimmer 30B & 73.66 & 77.70 & +4.04 & 8 \\
Kimi K3 & 72.99 & 86.72 & +13.73 & 4 \\
Qwen3.6-27B & 68.27 & 79.12 & +10.86 & 7 \\
Qwen3.5-122B-A10B & 53.32 & 68.04 & +14.72 & 10 \\
GLM-5.2 + GLM-5V-Turbo & 41.10 & withheld & -- & -- \\
\bottomrule
\end{tabular}
\end{table}

\begin{table}[t]
\caption{Equal-suite primary composite and task-weighted diagnostic. The point-
estimate ordering is unchanged for all eleven systems.}
\label{tab:ablweight}
\centering\footnotesize
\begin{tabular}{@{}lrr@{}}
\toprule
System & Equal suite & Task weighted \\
\midrule
GPT-5.6 Sol & 88.34 & 90.01 \\
Claude Fable 5 & 87.37 & 88.81 \\
GPT-5.6 Terra & 85.95 & 87.17 \\
Qwen3.8-27B & 82.54 & 84.15 \\
GPT-5.6 Luna & 78.65 & 81.97 \\
Claude Sonnet 5 & 75.40 & 78.17 \\
Muse Glimmer 30B & 73.66 & 76.64 \\
Kimi K3 & 72.99 & 75.30 \\
Qwen3.6-27B & 68.27 & 72.46 \\
Qwen3.5-122B-A10B & 53.32 & 55.51 \\
GLM-5.2 + GLM-5V-Turbo & 41.10 & 43.43 \\
\bottomrule
\end{tabular}
\end{table}

\subsection{Pairwise interval detail}

\begin{figure}[t]
\centering
\includegraphics[width=\textwidth]{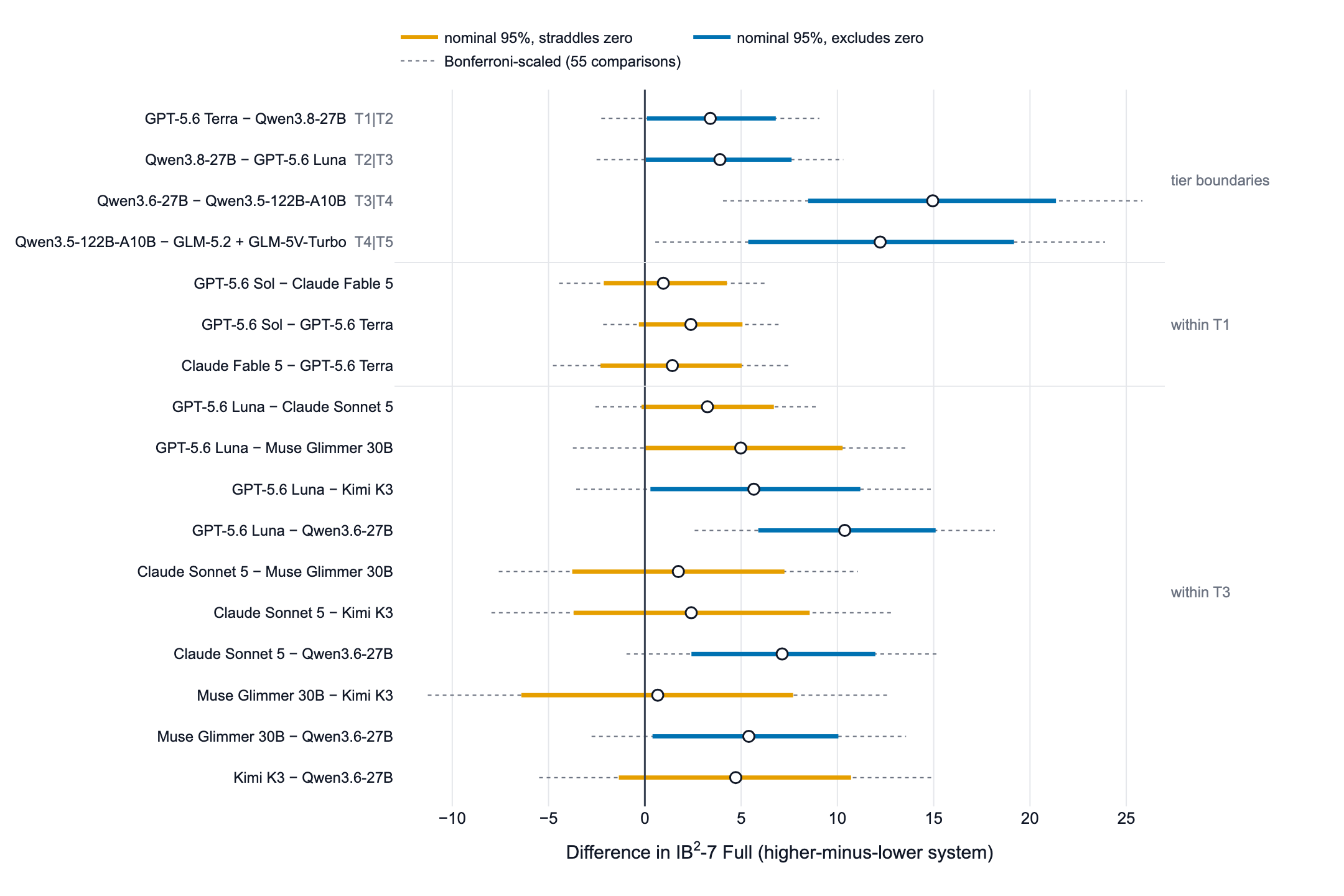}
\caption{The intervals the nominal tier construction rests on: the four tier-boundary
cuts, all three within-T1 pairs, and all ten within-T3 pairs, each stated as
higher-minus-lower so no sign convention is needed. Solid bars are the released
95\% percentile intervals of Algorithm~\ref{alg:bootstrap}, blue where they
exclude zero and orange where they straddle it; the open marker is the observed
difference. The dotted bar is a Bonferroni-scaled band at
$\alpha/55$. It is a \emph{normal-approximation rescaling} of the
released interval, not a recomputed adjusted percentile interval, because the
20{,}000 retained draws are held in the sealed run evidence rather than in
\texttt{pairwise\_intervals.csv}; it is a display of the multiplicity argument
in Section~\ref{sec:tiers}, not a second inferential claim. Both top cuts lose
zero exclusion under it while both lower cuts keep it, which is the
T1--T3 merge that paragraph describes.}
\label{fig:forest}
\end{figure}


\section{Synthetic scoring-conformance fixtures, one per suite}
\label{app:worked}

The private task bank and gold remain sealed. To make the scoring procedure
inspectable without leaking bank content, this appendix supplies seven complete
\emph{synthetic conformance fixtures}, one per suite. They were authored after
the reported runs, contain no private task text or artifact values, and use
synthetic responses rather than evaluated-system outputs. They test whether an
implementation of the public protocol scores a transparent case as specified;
they do not validate the difficulty, representativeness, or item quality of the
private bank. In each fixture the excerpt shown below is the complete synthetic
input relevant to the prompt. Their point allocations are likewise
illustrative: they are chosen to make one predicate legible per fixture and are
not instances of the suite-family weighting rule of
Section~\ref{sec:scoring}, which governs the sealed bank and is stated there in
full.

Three fixtures pass cleanly and four fail, so that partial credit, the numeric
tolerance predicate, sign handling, tool-trace ordering, and citation provenance
are each exercised rather than merely described. A fixture that always scores
100 demonstrates only that the scorer can add. The realized totals are S1 100,
S2 100, S3 55, S4 100, S5 65, S6 45, and S7 85.

\paragraph{S1: single-tab spreadsheet reasoning.}
\begin{quote}\small
\textbf{Prompt.} ``Using the Orders sheet, include only Q4 rows with status
Shipped or Returned. Revenue is units $\times$ price $\times(1-discount)$;
Returned rows are negative. Return net revenue, the highest-net-revenue region,
that region's amount, and a sheet-range citation.''

\textbf{Artifact.} \texttt{Orders!A1:F5}: headings
\{date,region,status,units,price,discount\}; rows
\{2026-10-04,West,Shipped,4,100,.10\},
\{2026-10-09,East,Returned,1,200,0\},
\{2026-11-01,West,Shipped,2,80,0\}, and
\{2026-12-03,East,Pending,9,50,0\}.

\textbf{Contract.} \texttt{\{net\_revenue:number, top\_region:string,
top\_region\_net:number, citations:array\}}.
\textbf{Gold.} A1 numeric 320.00, 40 points, absolute tolerance .01; A2 exact
``West'', 20; A3 numeric 520.00, 30, tolerance .01; A4 citation contains
\texttt{Orders!A1:F5}, 10.
\textbf{Synthetic response.} 320.00; West; 520.00;
\texttt{Orders!A1:F5}. \textbf{Score.} A1 40/40, A2 20/20, A3 30/30,
A4 10/10; total 100/100.
\end{quote}

\paragraph{S2: multi-tab spreadsheet join.}
\begin{quote}\small
\textbf{Prompt.} ``Join Sales to Products on product\_id. For channel Direct,
return recognized margin $\sum quantity\times(unit\_price-unit\_cost)$ and the
product with the largest contribution. Cite both ranges.''
\textbf{Artifacts.} \texttt{Sales!A1:D4} has
\{S1,P1,Direct,3\}, \{S2,P2,Partner,8\}, \{S3,P2,Direct,2\} under
\{sale\_id,product\_id,channel,quantity\}. \texttt{Products!A1:C3} has
\{P1,50,20\}, \{P2,80,55\} under \{product\_id,unit\_price,unit\_cost\}.
\textbf{Contract.} \texttt{\{margin:number, top\_product:string,
citations:array\}}. \textbf{Gold.} A1 numeric 140, 50 points, tolerance .01;
A2 exact P1, 30; A3 citations contain both displayed ranges, 20.
\textbf{Synthetic response.} 140; P1; both ranges. \textbf{Score.} 50/50,
30/30, 20/20; total 100/100.
\end{quote}

\paragraph{S3: selectable-table PDF analysis.}
\begin{quote}\small
\textbf{Prompt.} ``The packet contains draft and certified tables. From the
latest certified table, report total remediation spend and the largest control
category. Cite page and table.'' \textbf{Artifact.} Page 2, ``Draft A'', lists
Access 4.0 and Change 5.0 million. Page 6, ``Certified C---issued 2026-03-01'',
lists Access 4.4, Change 5.1, and Resilience 2.9 million.
\textbf{Contract.} \texttt{\{total\_million:number,
largest\_category:string,citations:array\}}. \textbf{Gold.} A1 numeric 12.4,
45 points, tolerance .01; A2 exact Change, 25; A3 citation matches page 6 and
table C, 30. \textbf{Synthetic response.} 12.45; Change; page 6, Table C.
\textbf{Score.} A1 0/45: $|12.45-12.4| = .05$ exceeds the .01 absolute
tolerance, and the numeric predicate is binary, so a near miss earns nothing;
A2 25/25, A3 30/30; total 55/100. The response read the correct table and
selected the correct category, and still scores below half, because the
reported figure is the deliverable.
\end{quote}

\paragraph{S4: image-table OCR.}
\begin{quote}\small
\textbf{Prompt.} ``Read the scanned invoice table. Parentheses denote credits.
Return the invoice total and the line with the largest absolute amount; cite the
image bounding box.'' \textbf{Artifact.} Synthetic image \texttt{invoice.png},
table box $[40,80,920,610]$: Hosting 1,250.00; Support 375.50; Credit
(125.00). \textbf{Contract.} \texttt{\{total:number,
largest\_line:string,citations:array\}}. \textbf{Gold.} A1 numeric 1500.50,
50 points, tolerance .01; A2 exact Hosting, 25; A3 image citation overlaps the
declared table box by at least .8 IoU, 25. \textbf{Synthetic response.}
1500.50; Hosting; \texttt{invoice.png [40,80,920,610]}.
\textbf{Score.} 50/50, 25/25, 25/25; total 100/100.
\end{quote}

\paragraph{S5: chart and structured-diagram interpretation.}
\begin{quote}\small
\textbf{Prompt.} ``From the variance chart, return the region with the largest
absolute plan variance, its signed percentage-point value, and whether it is
above or below plan. Cite the chart.'' \textbf{Artifact.} Synthetic horizontal
bars: North $+7$, South $-4$, East $-12$, West $+5$ percentage points; chart
title ``Q2 Plan Variance''. \textbf{Contract.}
\texttt{\{region:string,variance\_pp:number,direction:enum,citations:array\}}.
\textbf{Gold.} A1 exact East, 30; A2 numeric $-12$, 35, tolerance 0; A3 exact
below, 20; A4 chart citation contains the title, 15.
\textbf{Synthetic response.} East; $12$; below; Q2 Plan Variance.
\textbf{Score.} A1 30/30; A2 0/35: the gold is $-12$ at tolerance 0 and the
response reports the magnitude without the sign; A3 20/20; A4 15/15; total
65/100. A3 and A2 disagree in the response itself, since ``below'' and
$+12$ cannot both hold. The scorer does not reconcile them: each assertion is
evaluated independently against gold, so an internally inconsistent answer
collects credit for the clause that happens to be right.
\end{quote}

\paragraph{S6: tool calling and workflow execution.}
\begin{quote}\small
\textbf{Prompt.} ``For account A17, open a Priority-2 case only if its open
incident is older than the 24-hour SLA. Return the account, incident age, and
case id.'' \textbf{Tools.} \texttt{crm\_get\_account(account\_id)} returns
\texttt{\{open\_incident:I9,age\_hours:31\}} for A17;
\texttt{crm\_open\_case(account\_id,incident\_id,priority)} returns C44.
\textbf{Contract.} \texttt{\{account\_id:string,age\_hours:number,
case\_id:string\}}. \textbf{Gold.} A1 ordered tool trace
\texttt{get(A17)$\to$open(A17,I9,P2)}, 55 points; A2 exact A17, 10; A3
numeric 31, 10, tolerance 0; A4 exact C44, 15; A5 no unnecessary tool, 10.
\textbf{Synthetic response.} The model calls
\texttt{crm\_open\_case(A17,I9,P2)} first and
\texttt{crm\_get\_account(A17)} second, then returns
\texttt{\{A17,31,C44\}}. \textbf{Score.} A1 0/55: the trace is
\texttt{open$\to$get}, not the required \texttt{get$\to$open}; A2 10/10, A3
10/10, A4 15/15, A5 10/10; total 45/100. A5 still passes: no unnecessary tool
was called, and parsimony is scored independently of ordering. This is the
failure the suite exists to catch. The final JSON is correct in every field.
An evaluation that graded only the answer would score this 100. It would miss
that the system opened a case before establishing that the SLA had been
breached.
\end{quote}

\paragraph{S7: governed database plus document intelligence.}
\begin{quote}\small
\textbf{Prompt.} ``Use the certified dictionary, not the legacy snapshot, to
find the current net-amount field. Sum it for 2026 Q2 and count rows above the
policy exception threshold. Return the query id and policy citation.''
\textbf{Artifacts.} Dictionary rows say \texttt{net\_amount\_v2}=superseded and
\texttt{net\_amount\_v3}=certified. The synthetic table has Q2 values 120, 95,
and 90 in \texttt{net\_amount\_v3}. Policy page 4 defines exception as amount
$>100$. Query \texttt{Q-17} returns sum 305 and count 1.
\textbf{Contract.} \texttt{\{total:number,exception\_count:integer,
query\_id:string,citations:array\}}. \textbf{Gold.} A1 SQL trace discovers the
dictionary and uses only \texttt{net\_amount\_v3}, 35 points; A2 numeric 305,
25, tolerance .01; A3 exact integer 1, 15; A4 exact Q-17, 10; A5 citation
matches policy page 4, 15. \textbf{Synthetic response.} 305; 1; Q-17; page 3, with the certified-field
query trace. \textbf{Score.} A1 35/35: the trace discovers the dictionary and
reads only \texttt{net\_amount\_v3}; A2 25/25, A3 15/15, A4 10/10; A5
0/15: the exception threshold is defined on policy page 4 and the response
cites page 3; total 85/100. Every computed value is correct and the governed
query is clean; the response loses only the provenance for the rule it applied,
which is the part an auditor would need.
\end{quote}


\section{Run-plan manifest, hashing, and price accounting}
\label{app:manifest}

\begin{table}[!ht]
\caption{Frozen sampling parameters. ``def.'' means the provider default was
used and not exposed as a controllable value; it is not an inferred numeric
setting. Frequency penalty is def. and seed is unsupported or unexposed for
every row. Every row requested 65{,}536 output tokens. Kimi K3 uses
top-$p=1.0$ on agentic S6/S7 turns and 0.95 otherwise.}
\label{tab:sampling}
\centering\scriptsize
\setlength{\tabcolsep}{5pt}
\begin{tabular}{@{}lrrrrrr@{}}
\toprule
System & Effort & Temp. & top-$p$ & top-$k$ & Presence & Repetition \\
\midrule
GPT-5.6 Sol & xhigh & def. & def. & def. & def. & def. \\
GPT-5.6 Terra & xhigh & def. & def. & def. & def. & def. \\
GPT-5.6 Luna & xhigh & def. & def. & def. & def. & def. \\
Claude Fable 5 & high & def. & def. & def. & def. & def. \\
Claude Sonnet 5 & high & def. & def. & def. & def. & def. \\
Qwen3.5-122B-A10B & high & def. & def. & def. & def. & def. \\
Qwen3.6-27B & high & 1.0 & .95 & 20 & 0.0 & 1.0 \\
Qwen3.8-27B & xhigh & 1.0 & .95 & 20 & 0.0 & 1.0 \\
Kimi K3 & max & 1.0 & .95$^{a}$ & def. & def. & def. \\
GLM-5.2 & xhigh & def. & def. & def. & def. & def. \\
GLM-5V-Turbo & high & def. & def. & def. & def. & def. \\
Muse Glimmer 30B & xhigh & 1.0 & .95 & 64 & def. & def. \\
\bottomrule
\end{tabular}
\end{table}

The immutable checkpoint has schema
\texttt{ibib-run-plan-v1}. Its \texttt{plan} object contains, field for field:
\texttt{run\_id}, \texttt{benchmark\_version}, \texttt{scorer\_version},
\texttt{profile}, \texttt{eligibility}, \texttt{corpus\_manifest\_path},
\texttt{corpus\_manifest\_sha256}, \texttt{tasks\_path},
\texttt{tasks\_file\_sha256}, \texttt{gold\_path},
\texttt{gold\_file\_sha256}, \texttt{task\_count}, \texttt{task\_ids},
\texttt{task\_set\_sha256}, \texttt{task\_order\_sha256},
\texttt{public\_task\_sha256\_by\_id},
\texttt{public\_request\_sha256\_by\_id},
\texttt{public\_request\_hash\_scope}, \texttt{unsupported\_suites},
\texttt{system}, \texttt{adapter}, \texttt{settings}, and
\texttt{harness\_contract}. Secrets in the adapter are replaced with the
literal \texttt{<redacted>} before hashing.

Let \textsc{Canon} be JSON serialization with UTF-8 encoding, keys sorted,
separators \texttt{,} and \texttt{:} with no added whitespace, and
\texttt{ensure\_ascii=false}. Then
\[
\texttt{run\_plan\_sha256}=\operatorname{SHA256}(\textsc{Canon}(\texttt{plan})).
\]
The checkpoint's \texttt{schema\_version} and \texttt{created\_at} wrap this
object; creation time is deliberately outside the plan hash. Task-set hashes
apply the same construction to sorted task identifiers, task-order hashes to
the ordered list, and each public-task hash to
\texttt{\{hash\_scope,task\}}. Each public-request hash is computed by the
adapter over the provider-visible task plus the resolved harness contract.
Algorithm~\ref{alg:resume}'s $\kappa$ is the single
\texttt{run\_plan\_sha256}; equality therefore binds every field above rather
than an independently maintained subset.


\subsection{Run-accounting snapshot and cost accounting}
\label{app:cost}

The signed report binds accounting snapshot revision 7, assembled at
\texttt{2026-08-22T05:27:47.683705+00:00}, with SHA-256
\texttt{81f4259109d3147b3743f6b9decc5b28a66f4237e89acd9cd0328c2d3d33880e}.
The assembly timestamp is not a claim that every listed rate was current on
that date: the snapshot preserves the historical rates used to account for the
retained runs, while routes with provider-returned cost use that envelope.
Table~\ref{tab:prices} reports USD per million tokens for families priced from
token partitions. The remaining families use provider-reported per-response
cost and therefore have no paper-invented token rate.

\begin{table}[!ht]
\caption{Frozen run-accounting snapshot. CW5/CW1 are five-minute and one-hour cache
write rates. ``Envelope'' means exact provider-returned per-response cost.}
\label{tab:prices}
\centering\scriptsize
\begin{tabular}{@{}lrrrrrl@{}}
\toprule
Family/model & Input & Cached input & CW5 & CW1 & Output & Basis \\
\midrule
GPT-5.6 Sol   & 5.00 & 0.50 & 6.25 & -- & 30.00 & token partitions \\
GPT-5.6 Terra & 2.50 & 0.25 & 3.125 & -- & 15.00 & token partitions \\
GPT-5.6 Luna  & 1.00 & 0.10 & 1.25 & -- & 6.00 & token partitions \\
Claude Fable 5 & 10.00 & 1.00 & 12.50 & 20.00 & 50.00 & Messages usage \\
Claude Sonnet 5 & 2.00 & 0.20 & 2.50 & 4.00 & 10.00 & Messages usage \\
OpenRouter routes & -- & -- & -- & -- & -- & \texttt{usage.cost} envelope \\
Muse Glimmer 30B / DeepInfra & -- & -- & -- & -- & -- & \texttt{usage.estimated\_cost} envelope \\
\bottomrule
\end{tabular}
\end{table}

For GPT-5.6, uncached input, cached reads, cache writes and output are billed as
separate partitions; cache writes use $1.25\times$ uncached input. If a
request exceeds 272{,}000 input tokens, the entire request uses $2\times$ input and
$1.5\times$ output rates. For Claude, input, five-minute cache creation,
one-hour cache creation, cache reads and output are separate partitions; the
cache-write multipliers are $1.25\times$ and $2\times$. For every OpenRouter
response, including each tool turn, the ledger sums \texttt{usage.cost} and
retains \texttt{cost\_details} and the exact upstream provider. For every direct
DeepInfra response it sums \texttt{usage.estimated\_cost}. Missing or
inconsistent provider cost telemetry holds the affected cost analysis; no
fallback estimate is publication-eligible.


\section{Agentic-benchmark best-practice conformance}
\label{app:abc}

Table~\ref{tab:abc} applies the Agentic Benchmark Checklist (ABC)
\citep{zhu2025abc} item by item. ``Partial'' means the mechanism exists but the
release evidence is incomplete; ``N/A'' means ABC ties the item to an outcome
or grader type that \ibib{} does not use. This is a conformance disclosure, not
a claim that ABC certification exists.

\begingroup
\footnotesize
\setlength{\LTleft}{0pt}
\setlength{\LTright}{0pt}
\begin{longtable}{@{}p{.75cm}p{1.05cm}p{9.0cm}@{}}
\caption{Item-level ABC conformance. T: task validity; O: outcome validity;
R: reporting.}\label{tab:abc}\\
\toprule
Item & Status & Evidence or reason \\
\midrule
\endfirsthead
\toprule
Item & Status & Evidence or reason \\
\midrule
\endhead
T.1 & Partial & Tool catalogs and evaluator code are content-addressed, but the public artifact does not yet enumerate every package/runtime version. \\
T.2 & Partial & Binding verifies live API accessibility before release; transient outages still occurred and remain measured. \\
T.3 & Yes & Algorithms~\ref{alg:bind}--\ref{alg:resume} fail closed, retain errors, cap recovery and prohibit silent substitution. \\
T.4 & Yes & S6 replays each task in an evaluator-owned deterministic sandbox; task state is keyed to the task and reconstructed for each run. \\
T.5 & Yes & Provider-visible tasks and requests are separately hashed; gold, graders and scores are excluded from provider and adjudication projections. \\
T.6 & Partial & Corpus, harness and run plan are immutable; hosted provider internals are not frozen and are therefore part of the named system under test. \\
T.7 & Partial & Automated reference conformance and deterministic regrading pass, but independent two-person gold review remains a frozen-v1 requirement. \\
T.8 & Yes & Every admitted task passes reference conformance and task/gold consistency before sealing. \\
T.9 & Yes & The ineligible reference adapter is the automated oracle and covers all seven suites. \\
T.10 & Partial & Schema, lock, trace and no-gold-leak tests target shortcut paths; no independent adversarial implementation audit has been completed. \\
\midrule
O.a--O.f & N/A & Fourteen items (O.a.1--2, O.b.1--3, O.c.1--2, O.d.1--2, O.e.1--3, O.f.1--2) are scoped by ABC to grader types \ibib{} does not use: whole-string and substring response graders, LLM-as-judge, and unit- or fuzz-tested generated code. \ibib{} grades typed deterministic assertions and governed tool traces by deterministic replay, so none of the fourteen has a \ibib{} referent. \\
O.g.1 & Yes & S6 gold specifies the permitted successful terminal state and required ordered calls for each sealed workflow. \\
O.g.2 & Partial & Trace policy checks required calls, arguments, order and side effects; the state space has not been independently audited for every irrelevant field. \\
O.g.3 & Yes & Multi-step state, recovery and audit requirements prevent a trivial terminal-state mutation from earning full credit. \\
O.h.1 & Yes & Every task carries a typed output contract; the provider wire contract and deterministic parser are published. \\
O.h.2 & Yes & Multi-field assertions, citations and execution traces make random full-credit success negligible. \\
O.I.1 & Partial & Weighted assertions and trace policies limit reward hacking, but no independent adversarial reward audit has been performed. \\
\midrule
R.1 & Partial & Protocol, schemas, aggregates and verification tooling are public; the sealed corpus and gold remain private. \\
R.2 & Partial & Provider-neutral interfaces and analysis code are public, but the controlled private evaluator is not yet deployed for third-party use. \\
R.3 & Yes & The scored bank is private and content-addressed; providers receive only the current task request. \\
R.4 & Yes & Versioning, reserve and anti-ceiling policies exist; the 2026-08-31 review records trigger evidence and pre-v1 replacement decisions. \\
R.5 & Yes & Table~\ref{tab:suites} states represented activity, setting, work product and evaluated result; Section~\ref{sec:saturation} limits construct claims. \\
R.6 & Yes & Definition~\ref{def:sut} defines the evaluated object as the complete served system, including route and harness. \\
R.7 & Yes & Locks, binding, score-blind adjudication, exact resume and complete-cohort resealing are explicit flaw controls. \\
R.8 & Yes & Section~\ref{sec:threats} and the ethics section discuss private-data, saturation, route, weighting and conflict limitations. \\
R.9 & Partial & Four protocol ablations quantify major choices; gold-label noise and run-to-run variance are not estimated. \\
R.10 & Yes & All 55 pairwise comparisons and the route delta use predeclared
20{,}000-replicate paired task-set sensitivity intervals. \\
R.11 & Yes & Coverage, first-pass, transport-corrected and conditional-accuracy layers are separated, with explicit withholding rules. \\
R.12 & No & No human or other non-AI baseline was run; procurement-grade interpretation is explicitly withheld. \\
R.13 & No & No do-nothing or random trivial agent was run. \\
\bottomrule
\end{longtable}
\endgroup

The explicit non-conformances are R.12--R.13. Material partials are T.7,
T.10/O.I.1, R.2, and R.9; Table~\ref{tab:abc} records the submission-relevant
limitations rather than converting them to passes.

\end{document}